\documentclass[10pt,twocolumn,letterpaper]{article}

\usepackage[pagenumbers]{cvpr} %
\usepackage[accsupp]{axessibility}

\definecolor{cvprblue}{rgb}{0.21,0.49,0.74}
\usepackage[pagebackref,breaklinks,colorlinks,allcolors=cvprblue]{hyperref}

\usepackage{pifont}
\usepackage{multirow}
\usepackage{standalone}
\usepackage{etoolbox}
\usepackage{tikz}
\usepackage{pgfplots}
\usepackage{pgfplotstable}
\usepackage{stfloats}
\pgfplotsset{compat=newest}
\usepgfplotslibrary{external}
\AtBeginEnvironment{tikzpicture}{%
    \renewcommand{\AtTextUpperLeft}{} %
}

\title{LPOSS: Label Propagation Over Patches and Pixels\\ for Open-vocabulary Semantic Segmentation}

\author{
    Vladan Stojnić$^1$
    \and
    Yannis Kalantidis$^2$
    \and
    Jiří Matas$^1$
    \and
    Giorgos Tolias$^1$ \and \\
    $^1$~{VRG, FEE, Czech Technical University in Prague} \hspace{1cm} $^2$~{NAVER LABS Europe}
}

\begin{document}
\maketitle
\newcommand{\mypartight}[1]{\noindent {\bf #1}}
\newcommand{\myparagraph}[1]{\vspace{3pt}\noindent\textbf{#1}\xspace}

\newcommand{\gt}[1]{{\color{purple}{GT: #1}}}
\newcommand{\gtt}[1]{{\color{purple}{#1}}}
\newcommand{\gtr}[2]{{\color{purple}\st{#1} {#2}}}
\newcommand{\vsc}[1]{{\color{blue}{VS: #1}}}
\newcommand{\vst}[1]{{\color{blue}{#1}}}
\newcommand{\yk}[1]{{\color{OliveGreen}{Y: #1}}}
\newcommand{\ykt}[1]{{\color{OliveGreen}{#1}}}
\newcommand{\cready}[1]{{\color{RoyalPurple}{#1}}}

\newcommand{\question}[2]{\noindent\textbf{#1} (#2)}

\newcommand{\gray}[1]{{\color{gray}{#1}}}
\newcommand{\alert}[1]{{\color{red}{#1}}}

\newcommand{\gain}[1]{\textbf{\color{OliveGreen}{{\scriptsize$\uparrow$}{\scriptsize #1}}}}
\newcommand{\loss}[1]{\textbf{\color{red}{{\scriptsize$\downarrow$}{\scriptsize #1}}}}

\newcommand{\rvi}[1]{{\color{NavyBlue}{\textbf{#1}}}}
\newcommand{\rvii}[1]{{\color{OliveGreen}{\textbf{#1}}}}
\newcommand{\rviii}[1]{{\color{BrickRed}{\textbf{#1}}}}

\newcommand{\rone}[0]{{\rvi{R-Kntg}\xspace}}
\newcommand{\rtwo}[0]{{\rvii{R-uTeG}\xspace}}
\newcommand{\rthree}[0]{{\rviii{R-aUnL}\xspace}}

\newcommand{\questionone}[1]{\noindent\emph{\textbf{\rvi{#1}}}}
\newcommand{\questiontwo}[1]{\noindent\emph{\textbf{\rvii{#1}}}}
\newcommand{\questionthree}[1]{\noindent\emph{\textbf{\rviii{#1}}}}

\def\roxf{$\mathcal{R}$Oxford\xspace}
\def\rox{$\mathcal{R}$Oxf\xspace}
\def\ro{$\mathcal{R}$O\xspace}
\def\rpar{$\mathcal{R}$Paris\xspace}
\def\rpa{$\mathcal{R}$Par\xspace}
\def\rp{$\mathcal{R}$P\xspace}
\def\rdis{$\mathcal{R}$1M\xspace}

\newcommand\resnet[3]{\ensuremath{\prescript{#2}{}{\mathtt{R}}{#1}_{\scriptscriptstyle #3}}\xspace}

\newcommand{\clipc}{CornflowerBlue}
\newcommand{\oursc}{MidnightBlue}
\newcommand{\ourssparsec}{MidnightBlue!50!Black}

\newcommand{\inmapc}{RedOrange}
\newcommand{\oursinmapc}{BrickRed}
\newcommand{\oursinmapsparsec}{BrickRed!50!Black}

\newcommand{\clipdnc}{OliveGreen}
\newcommand{\tptc}{DarkOrchid}
\newcommand{\ours}{LPOSS\xspace} %
\newcommand{\oursplus}{LPOSS+\xspace} %

\newcommand{\stddev}[1]{\scriptsize{$\pm#1$}}

\newcommand{\diffup}[1]{{\color{OliveGreen}{($\uparrow$ #1)}}}
\newcommand{\diffdown}[1]{{\color{BrickRed}{($\downarrow$ #1)}}}

\crefname{section}{Sec.}{Secs.}
\Crefname{section}{Sec.}{Secs.}
\Crefname{table}{Tab.}{Tabs.}
\crefname{table}{Tab.}{Tabs.}
\Crefname{figure}{Fig.}{Figs.}
\crefname{figure}{Fig.}{Figs.}
\Crefname{equation}{Eq.}{Eqs.}
\crefname{equation}{Eq.}{Eqs.}
\crefname{appendix}{Appendix}{Appendices}

\newcommand{\comment} [1]{{\color{orange} \Comment     #1}} %

\def\nmsp{\hspace{-6pt}}
\def\nssp{\hspace{-3pt}}
\def\nxssp{\hspace{-1pt}}
\def\zsp{\hspace{0pt}}
\def\xssp{\hspace{1pt}}
\def\ssp{\hspace{3pt}}
\def\msp{\hspace{6pt}}
\def\lsp{\hspace{12pt}}
\def\xlsp{\hspace{20pt}}

\newcommand{\head}[1]{{\smallskip\noindent\bf #1}}
\newcommand{\equ}[1]{(\ref{equ:#1})\xspace}

\def\Linv{\ensuremath{L_{\text{inv}}}\xspace}

\newcommand{\nn}[1]{\ensuremath{\text{NN}_{#1}}\xspace}
\newcommand{\knn}{\ensuremath{\text{kNN}}\xspace}
\def\l1{\ensuremath{\ell_1}\xspace}
\def\l2{\ensuremath{\ell_2}\xspace}

\newcommand{\tran}{^\top}
\newcommand{\mtran}{^{-\top}}
\newcommand{\zcol}{\mathbf{0}}
\newcommand{\zrow}{\zcol\tran}

\newcommand{\ind}{\mathds{1}}
\newcommand{\expect}{\mathbb{E}}
\newcommand{\nat}{\mathbb{N}}
\newcommand{\zahl}{\mathbb{Z}}
\newcommand{\real}{\mathbb{R}}
\newcommand{\proj}{\mathbb{P}}
\newcommand{\prob}{\mathbf{Pr}}

\newcommand{\mif}{\textrm{if }}
\newcommand{\other}{\textrm{otherwise}}
\newcommand{\minimize}{\textrm{minimize }}
\newcommand{\maximize}{\textrm{maximize }}

\newcommand{\id}{\operatorname{id}}
\newcommand{\const}{\operatorname{const}}
\newcommand{\sgn}{\operatorname{sgn}}
\newcommand{\var}{\operatorname{Var}}
\newcommand{\mean}{\operatorname{mean}}
\newcommand{\trace}{\operatorname{tr}}
\newcommand{\diag}{\operatorname{diag}}
\newcommand{\vect}{\operatorname{vec}}
\newcommand{\cov}{\operatorname{cov}}
\newcommand{\argmax}{\operatorname{argmax}}

\newcommand{\softmax}{\operatorname{softmax}}
\newcommand{\clip}{\operatorname{clip}}

\newcommand{\defn}{\mathrel{:=}}
\newcommand{\peq}{\mathrel{+\!=}}
\newcommand{\meq}{\mathrel{-\!=}}

\newcommand{\floor}[1]{\left\lfloor{#1}\right\rfloor}
\newcommand{\ceil}[1]{\left\lceil{#1}\right\rceil}
\newcommand{\inner}[1]{\left\langle{#1}\right\rangle}
\newcommand{\norm}[1]{\left\|{#1}\right\|}
\newcommand{\frob}[1]{\norm{#1}_F}
\newcommand{\card}[1]{\left|{#1}\right|\xspace}
\newcommand{\diff}{\mathrm{d}}
\newcommand{\der}[3][]{\frac{d^{#1}#2}{d#3^{#1}}}
\newcommand{\pder}[3][]{\frac{\partial^{#1}{#2}}{\partial{#3^{#1}}}}
\newcommand{\ipder}[3][]{\partial^{#1}{#2}/\partial{#3^{#1}}}
\newcommand{\dder}[3]{\frac{\partial^2{#1}}{\partial{#2}\partial{#3}}}

\newcommand{\wb}[1]{\overline{#1}}
\newcommand{\wt}[1]{\widetilde{#1}}

\newcommand{\cA}{\mathcal{A}}
\newcommand{\cB}{\mathcal{B}}
\newcommand{\cC}{\mathcal{C}}
\newcommand{\cD}{\mathcal{D}}
\newcommand{\cE}{\mathcal{E}}
\newcommand{\cF}{\mathcal{F}}
\newcommand{\cG}{\mathcal{G}}
\newcommand{\cH}{\mathcal{H}}
\newcommand{\cI}{\mathcal{I}}
\newcommand{\cJ}{\mathcal{J}}
\newcommand{\cK}{\mathcal{K}}
\newcommand{\cL}{\mathcal{L}}
\newcommand{\cM}{\mathcal{M}}
\newcommand{\cN}{\mathcal{N}}
\newcommand{\cO}{\mathcal{O}}
\newcommand{\cP}{\mathcal{P}}
\newcommand{\cQ}{\mathcal{Q}}
\newcommand{\cR}{\mathcal{R}}
\newcommand{\cS}{\mathcal{S}}
\newcommand{\cT}{\mathcal{T}}
\newcommand{\cU}{\mathcal{U}}
\newcommand{\cV}{\mathcal{V}}
\newcommand{\cW}{\mathcal{W}}
\newcommand{\cX}{\mathcal{X}}
\newcommand{\cY}{\mathcal{Y}}
\newcommand{\cZ}{\mathcal{Z}}

\newcommand{\vA}{\mathbf{A}}
\newcommand{\vB}{\mathbf{B}}
\newcommand{\vC}{\mathbf{C}}
\newcommand{\vD}{\mathbf{D}}
\newcommand{\vE}{\mathbf{E}}
\newcommand{\vF}{\mathbf{F}}
\newcommand{\vG}{\mathbf{G}}
\newcommand{\vH}{\mathbf{H}}
\newcommand{\vI}{\mathbf{I}}
\newcommand{\vJ}{\mathbf{J}}
\newcommand{\vK}{\mathbf{K}}
\newcommand{\vL}{\mathbf{L}}
\newcommand{\vM}{\mathbf{M}}
\newcommand{\vN}{\mathbf{N}}
\newcommand{\vO}{\mathbf{O}}
\newcommand{\vP}{\mathbf{P}}
\newcommand{\vQ}{\mathbf{Q}}
\newcommand{\vR}{\mathbf{R}}
\newcommand{\vS}{\mathbf{S}}
\newcommand{\vT}{\mathbf{T}}
\newcommand{\vU}{\mathbf{U}}
\newcommand{\vV}{\mathbf{V}}
\newcommand{\vW}{\mathbf{W}}
\newcommand{\vX}{\mathbf{X}}
\newcommand{\vY}{\mathbf{Y}}
\newcommand{\vZ}{\mathbf{Z}}

\newcommand{\va}{\mathbf{a}}
\newcommand{\vb}{\mathbf{b}}
\newcommand{\vc}{\mathbf{c}}
\newcommand{\vd}{\mathbf{d}}
\newcommand{\ve}{\mathbf{e}}
\newcommand{\vf}{\mathbf{f}}
\newcommand{\vg}{\mathbf{g}}
\newcommand{\vh}{\mathbf{h}}
\newcommand{\vi}{\mathbf{i}}
\newcommand{\vj}{\mathbf{j}}
\newcommand{\vk}{\mathbf{k}}
\newcommand{\vl}{\mathbf{l}}
\newcommand{\vm}{\mathbf{m}}
\newcommand{\vn}{\mathbf{n}}
\newcommand{\vo}{\mathbf{o}}
\newcommand{\vp}{\mathbf{p}}
\newcommand{\vq}{\mathbf{q}}
\newcommand{\vr}{\mathbf{r}}
\newcommand{\Vs}{\mathbf{s}}
\newcommand{\vt}{\mathbf{t}}
\newcommand{\vu}{\mathbf{u}}
\newcommand{\vv}{\mathbf{v}}
\newcommand{\vw}{\mathbf{w}}
\newcommand{\vx}{\mathbf{x}}
\newcommand{\vy}{\mathbf{y}}
\newcommand{\vz}{\mathbf{z}}

\newcommand{\pp}{\mathtt{p}}

\newcommand{\vone}{\mathbf{1}}
\newcommand{\vzero}{\mathbf{0}}

\newcommand{\valpha}{{\boldsymbol{\alpha}}}
\newcommand{\vbeta}{{\boldsymbol{\beta}}}
\newcommand{\vgamma}{{\boldsymbol{\gamma}}}
\newcommand{\vdelta}{{\boldsymbol{\delta}}}
\newcommand{\vepsilon}{{\boldsymbol{\epsilon}}}
\newcommand{\vzeta}{{\boldsymbol{\zeta}}}
\newcommand{\veta}{{\boldsymbol{\eta}}}
\newcommand{\vtheta}{{\boldsymbol{\theta}}}
\newcommand{\viota}{{\boldsymbol{\iota}}}
\newcommand{\vkappa}{{\boldsymbol{\kappa}}}
\newcommand{\vlambda}{{\boldsymbol{\lambda}}}
\newcommand{\vmu}{{\boldsymbol{\mu}}}
\newcommand{\vnu}{{\boldsymbol{\nu}}}
\newcommand{\vxi}{{\boldsymbol{\xi}}}
\newcommand{\vomikron}{{\boldsymbol{\omikron}}}
\newcommand{\vpi}{{\boldsymbol{\pi}}}
\newcommand{\vrho}{{\boldsymbol{\rho}}}
\newcommand{\vsigma}{{\boldsymbol{\sigma}}}
\newcommand{\vtau}{{\boldsymbol{\tau}}}
\newcommand{\vupsilon}{{\boldsymbol{\upsilon}}}
\newcommand{\vphi}{{\boldsymbol{\phi}}}
\newcommand{\vchi}{{\boldsymbol{\chi}}}
\newcommand{\vpsi}{{\boldsymbol{\psi}}}
\newcommand{\vomega}{{\boldsymbol{\omega}}}

\newcommand{\rLambda}{\mathrm{\Lambda}}
\newcommand{\rSigma}{\mathrm{\Sigma}}

\makeatletter
\DeclareRobustCommand\onedot{\futurelet\@let@token\@onedot}
\def\@onedot{\ifx\@let@token.\else.\null\fi\xspace}
\def\eg{\emph{e.g}\onedot} \def\Eg{\emph{E.g}\onedot}
\def\ie{\emph{i.e}\onedot} \def\Ie{\emph{I.e}\onedot}
\def\vs{\emph{vs\onedot}}
\def\cf{\emph{cf}\onedot} \def\Cf{\emph{C.f}\onedot}
\def\etc{\emph{etc}\onedot} \def\vs{\emph{vs}\onedot}
\def\wrt{w.r.t\onedot} \def\dof{d.o.f\onedot}
\def\etal{\emph{et al}\onedot}
\makeatother

\begin{abstract}
We propose a training-free method for open-vocabulary semantic segmentation using Vision-and-Language Models (VLMs). Our approach enhances the initial per-patch predictions of VLMs through label propagation, which jointly optimizes predictions by incorporating patch-to-patch relationships. Since VLMs are primarily optimized for cross-modal alignment and not for intra-modal similarity, we use a Vision Model (VM) that is observed to better capture these relationships.
We address resolution limitations inherent to patch-based encoders by applying label propagation at the pixel level as a refinement step, significantly improving segmentation accuracy near class boundaries. Our method, called \oursplus, performs inference over the entire image, avoiding window-based processing and thereby capturing contextual interactions across the full image. \oursplus achieves state-of-the-art performance among training-free methods, across a diverse set of datasets. Code: \url{https://github.com/vladan-stojnic/LPOSS}
\end{abstract}

\section{Introduction}
\label{sec:intro}
Semantic segmentation models are typically trained in a supervised manner, requiring pixel-level annotations based on a predefined set of categories or classes.  
These approaches face significant scalability challenges due to the high cost of annotation, and their applicability is limited by the fixed set of classes they are trained to recognize.
Open-vocabulary semantic segmentation aims to generalize and allow the segmentation of an image into any given list of classes provided at inference time. This task is usually solved with the use of a Vision-and-Language Model (VLM)~\cite{radford21clip,jia21align,cherti23openclip}.

VLMs produce aligned textual and visual features and enable matching textual descriptions, such as class names, to visual features. While highly effective for image-level classification tasks, VLMs are less suited for dense prediction tasks like semantic segmentation out-of-the-box, as they are trained to align global image representations with textual data. Our work addresses this gap by proposing a novel, training-free VLM enhancement for open-vocabulary semantic segmentation. 

\begin{figure}
\vspace{-5pt}
\small
    \centering
    \begin{tabular}{@{\xssp}c@{\msp}c@{\msp}c@{\xssp}}
    & 53.8/35.9 & 66.1/48.4\\[-1pt]
    \includegraphics[width=.15\textwidth]{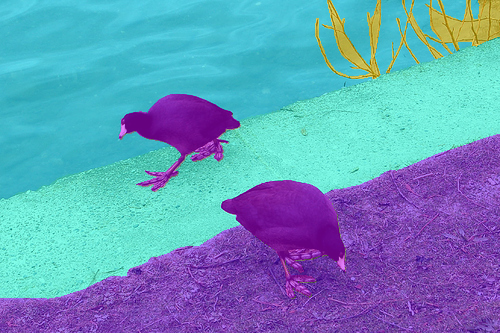}&
    \includegraphics[width=.15\textwidth]{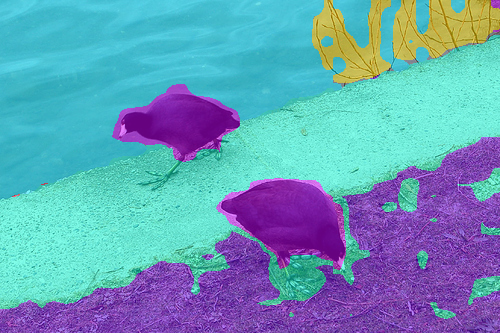}&
    \includegraphics[width=.15\textwidth]{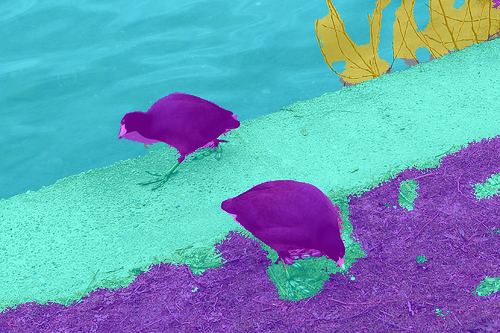}\\
    & 82.1/53.6 & 84.0/59.1\\[-1pt]
    \includegraphics[width=.15\textwidth]{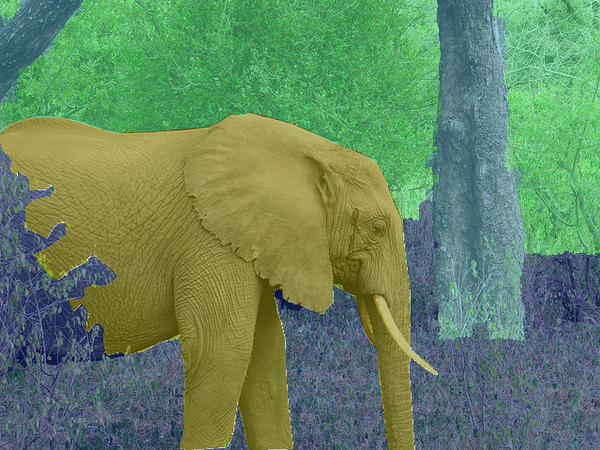}&
    \includegraphics[width=.15\textwidth]{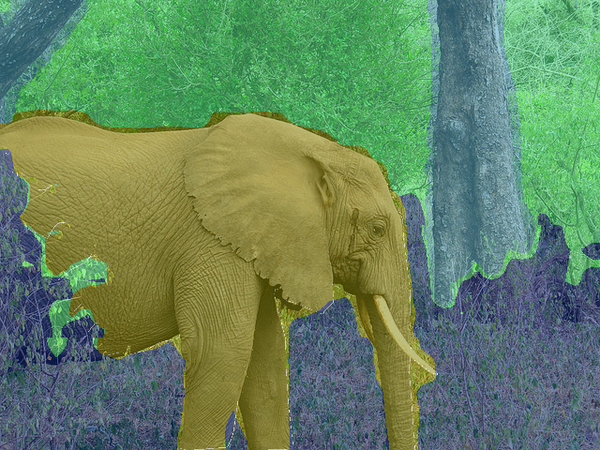}&
    \includegraphics[width=.15\textwidth]{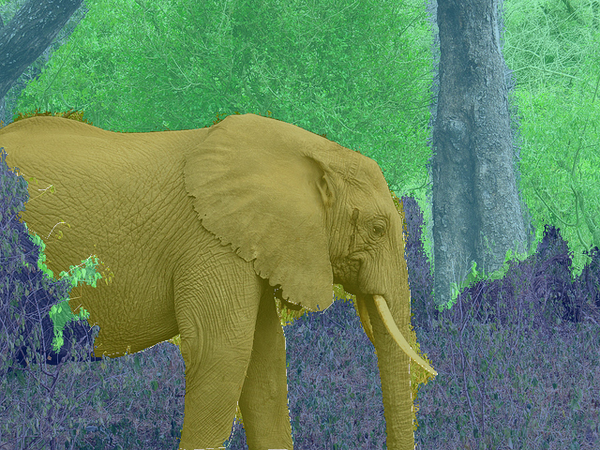}\\  
    & 69.4/54.5 & 70.3/60.0\\[-1pt]
    \includegraphics[width=.15\textwidth]{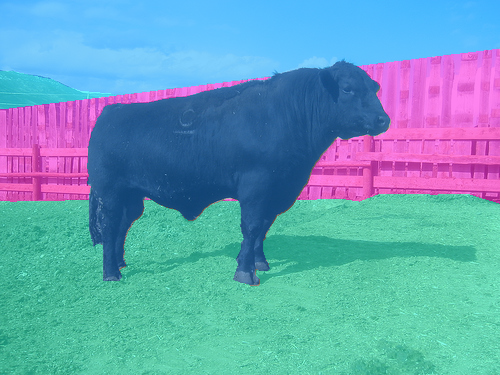}&
    \includegraphics[width=.15\textwidth]{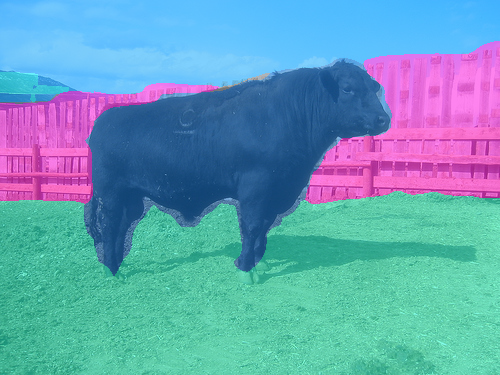}&
    \includegraphics[width=.15\textwidth]{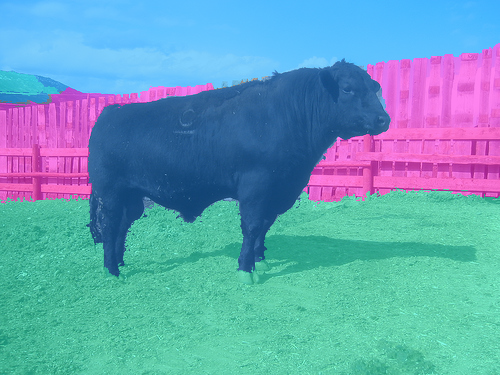}\\
    ground-truth & \ours & \oursplus\\
    \end{tabular}
    \caption{Open-vocabulary semantic segmentation with \ours (patch-level LP) and \oursplus (pixel-level LP). Performance reported via mIoU / Boundary IoU. Images from the Context and COCO-Stuff datasets.
    \vspace{-10pt}
    \label{fig:lposs_comparison}
    }
\end{figure}

Although segmentation is defined as pixel-wise labeling, recent approaches, that rely on VLMs~\cite{wang2024sclip,wysoczanska2024clip,lan2024proxyclip} utilizing ViT~\cite{dosovitskiy2021an} architecture, form predictions at the patch level and
simply up-sample the predictions. A ViT-based VLM typically provides a class probability distribution independently for each patch, yielding an initial prediction. Our approach incorporates the initial prediction together with patch relationships by making joint predictions across all patches. This is achieved through minimizing a quadratic cost function with unary and binary terms.
This approach ensures that final predictions remain close to initial ones while nearby patches or patches with similar appearances produce similar predictions, reflecting a common prior in semantic segmentation.
Optimization is performed through either an iterative Label Propagation (LP)~\cite{zhou03lp} operation on a patch graph or, equivalently, by solving a linear system. The latter is computationally efficient due to the high sparsity of the coefficient matrix and the use of GPU-based solvers. 
Following prior work~\cite{wysoczanska2024clip,kang2024defense,lan2024proxyclip}, and due to the fact that VLMs optimize inter-modal rather than intra-modal similarity, we leverage a strong Vision Model (VM), specifically DINO~\cite{caron21dino}, to effectively capture patch-to-patch appearance and relationships.
Using such an encoder, instead of the VLM’s visual encoder, significantly improves predictions at a patch level.

While operating at the patch level provides efficiency benefits, it constrains pixel-level accuracy. 
In an experiment where semantic segmentation ground truth 
is downsampled to patch-level resolution, then upsampled and evaluated as a prediction, we show that the resulting performance measured via mIoU averages approximately 85\% across eight datasets.
To address this limitation, we perform a second stage of LP on a pixel-level graph, using the patch-based LP result as the initial prediction set. This refinement consistently improves segmentation accuracy, particularly near class boundaries, as we show by measuring the Boundary IoU metric~\cite{cheng2021boundary}. Figure~\ref{fig:lposs_comparison} demonstrates  examples of such improvement using VLM and VM with a ViT backbone.

The strength of ViT architectures comes with limitations. 
There is a strong bias introduced by their pre-training settings, \ie processing fixed-resolution square images.
Consequently, current segmentation methods perform feature extraction and prediction in a sliding window fashion with overlapping windows. 
In contrast, we move beyond this standard practice by performing predictions jointly for the entire image, accounting for interactions across all patches. However, we show that while joint predictions are possible, joint feature extraction across the whole image is not feasible without a noticeable performance decrease.
Our proposed method, called \textit{Label Propagation over Patches} (\ours) \textit{and Pixels} (\oursplus), achieves state-of-the-art results across various datasets, outperforming existing training-free methods 
on eight datasets.
It also outperforms pixel-level VLM fine-tuning approaches, when the test distribution differs significantly from the training one.
Our contributions are summarized as follows:
\begin{itemize}
        \item We establish a connection between classical label propagation methods and vision-language models (VLMs) in the context of dense prediction tasks.

        \item We introduce a training-free approach for semantic segmentation that predicts and propagates labels at both patch and pixel levels. This strategy leverages the pre-training of foundation models and addresses the performance limitations of coarse patch-level predictions.
        
    \item We propose to decouple feature extraction from class prediction, enabling 
    image-level inference and capturing contextual interactions across the entire image.
\end{itemize}

\section{Related work}
\label{sec:related}

The introduction of vision-language models~\cite{cherti23openclip,jia21align,radford21clip} opened up the possibilities of performing many recognition tasks in the open-vocabulary setting~\cite{wu24openvocabsurvery}, \eg, zero-shot classification~\cite{radford21clip,stojnic24zlap,udandarao23susx}, semantic segmentation~\cite{zhou2022extract,ghiasi22openseg}, object detection~\cite{wu23baron}, \etc. However, considering these models are usually trained only to optimize global image representation, they do not excel in dense recognition tasks such as segmentation~\cite{zhou2022extract,bousselham24gem}. Because of this, considerable work is being done to improve their performance. This is achieved by training specialized VLM for such tasks, changing the VLM architecture during inference, and employing other vision models to aid VLMs.

\noindent\textbf{Training VLMs for segmentation.}
One line of work trains from scratch VLMs specialized for semantic segmentation using varying levels of supervision: using segmentation masks without class annotations~\cite{ding22zegformer,ghiasi22openseg} and only image-caption pairs~\cite{luo23segclip,ranasinghe23perceptual,xu22groupvit,xu23ovsegmentor}. On the other hand, another line of work trains additional modules on top of pre-trained VLM to improve their localization abilities~\cite{cha23tcl,liang23maskadapted,mukhoti23pacl}. 

\noindent\textbf{Training-free segmentation with VLMs.} It is observed~\cite{wang2024sclip,li22surgery} that different VLM components are responsible for VLMs bad localization performance and that the performance can be substantially improved by removing or changing them during inference. MaskCLIP~\cite{zhou2022extract} revisits the global attention pooling of CLIP~\cite{radford21clip} and proposes to remove its query and key embeddings while reformulating the value embedding layer and the last linear layer into two $1 \times 1$ convolutions. SCLIP~\cite{wang2024sclip} proposes to replace the self-attention layer of the ViT with correlative self-attention. ClearCLIP~\cite{lan2024clearclip} identifies residual connections as the primary source of VLMs' bad localization performance. Because of this, it proposes to remove them from the last layer. It additionally replaces self-attention with self-self-attention and removes the feed-forward network from the last layer. GEM~\cite{bousselham24gem} improves the semantic segmentation abilities of VLMs by adding self-self-attention blocks parallel to self-attention layers. We consider these methods complementary to ours as we apply our method on top of MaskCLIP~\cite{zhou2022extract}.

\noindent\textbf{VLMs and VMs for segmentation.} Vision models trained in a self-supervised fashion show remarkable object localization abilities~\cite{oquab2024dinov2,darcet24registers,caron21dino,simeoni21lost,simeoni23found}. Because of this, the growing line of work investigates how to leverage them to improve the ability of VLMs to perform open-vocabulary semantic segmentation~\cite{wysoczanska2024clip,lan2024proxyclip} or other segmentation tasks~\cite{ypsilantis24cbnc}. CLIP-DIY~\cite{wysoczanska2024diy} uses an unsupervised object discovery method FOUND~\cite{simeoni23found} to clear and improve the open-vocabulary segmentation maps of CLIP~\cite{radford21clip}. CLIP-DINOiser~\cite{wysoczanska2024clip} proposes a training-free variant that uses patch-to-patch affinity from DINO~\cite{caron21dino} to refine CLIP's image features. Additionally, it shows that good localization capabilities can be extracted from CLIP by training two small convolutional layers. ProxyCLIP~\cite{lan2024proxyclip} proposes replacing the attention matrix of the last layer of CLIP~\cite{radford21clip} with an affinity matrix from a vision model such as DINO~\cite{caron21dino}. LaVG performs panoptic segmentation using the normalized cut on top of DINO~\cite{caron21dino} features. The discovered segments are then classified into a class using a VLM. Our method falls into this group of methods and is most similar to CLIP-DINOiser~\cite{wysoczanska2024clip}. Compared to CLIP-DINOiser, which uses affinity based on Euclidean similarities between DINO patches, we propose to use geodesic similarities obtained through label propagation. Additionally, compared to all other works, we perform the prediction jointly across all sliding windows, which are usually used in such methods. Finally, we depart from performing predictions only at the patch level and refine the predictions at the pixel level by applying the same process on top of pixels.

\noindent\textbf{Relation to CRFs.}
Conditional Random Fields (CRFs)~\cite{sutton12crfbook,krahenbuhl11densecrf} have a long tradition in dense prediction tasks with classical approaches~\cite{he04multiscale,shotton06texton,krahenbuhl11densecrf}. 
In early deep models for semantic segmentation, CRFs are used as a refinement step to increase the accuracy and the resolution of the predictions~\cite{chen18deeplab}. 
There are also attempts to integrate a similar objective function during end-to-end training~\cite{zheng15crfrnn,chandra16fast}. 
CRFs and LP both leverage pairwise connections to achieve smooth label assignments. LP directly propagates labels based on neighbors, while CRFs explicitly model these relationships in their energy function.
We opt for an LP variant that has theoretical convergence proofs and corresponds to an intuitive quadratic cost function, which does not hold for all existing LP variants~\cite{bengio06semibook}. 
LP is typically easy to implement, computationally efficient~\cite{zhou03lp}, and compatible with GPU-based software implementations~\cite{stojnic24zlap}.
Consequently, it is used in various computer vision and machine learning tasks~\cite{stojnic24zlap,hu24reclip,sun24corrmatch,iscen19lpsemi}, while in this work, LP is tailored for semantic segmentation in the era of VLMs and ViT encoders.

\section{Label propagation over patches and pixels}
\label{sec:method}

\subsection{Problem formulation}
In the task of open-vocabulary semantic segmentation we are given an image $X \in \real^{H \times W \times 3}$ and a list of class names $\cC = \{c_1, \ldots, c_C\}$. Our goal is to assign one of the classes to every pixel or, conversely, tensor $Y \in \real^{H \times W \times C}$ that contains a segmentation mask for each of the $C$ classes.

To solve such a task, one can utilize a Vision Language Model (VLM), \ie a model that is able to encode both the class names as well as the image in the compatible feature spaces. Let $f: \cI \rightarrow \real^{d\times N_P }$ denote the vision encoder, \ie a ViT~\cite{dosovitskiy2021an} with patch size $P$. The vision encoder extracts feature vectors for all $N_P = \ceil{H/P} \cdot \ceil{W/P}$ patches for an image in $\cI$\footnote{We discard the representation corresponding to the CLS token usually present in ViTs.}.
Also, let $g: \cT \rightarrow \real^d$ represent the text encoder of the VLM,  with $\cT$ representing the space of the textual input. 
One can, therefore, perform open-vocabulary semantic segmentation by obtaining the patch features for an image $Z_{\text{vlm}} = f(X)$, class name features $F = g(\cC)\in \real^{d\times C}$, and calculating per patch class similarities 
\begin{equation}
Y_{\text{vlm}} = Z_{\text{vlm}}^\top F,
\label{equ:baseline}
\end{equation}
which are further upsampled to pixel-wise class similarities at the original image size. The class with the highest similarity is then assigned to each pixel.

\subsection{Preliminaries}
\label{sec:preliminaries}

\noindent\textbf{Label propagation (LP).}
Let $\{ n_{1}, \ldots, n_{N} \}$ be a set of graph nodes and $S \in \real^{N \times N}$ be a symmetric, typically very sparse, adjacency matrix with zero diagonal. We obtain a symmetrically normalized adjacency matrix by $\hat{S} = D^{-\frac{1}{2}} S D^{-\frac{1}{2}}$, where $D = \text{diag}(S\vone_N)$ is a degree matrix and $\vone_N$ is the all-ones $N$-dimensional vector. Given $\hat{S}$, label propagation~\cite{zhou03lp} is an iterative process given by
\begin{equation}
    \hat{Y}^{(t+1)} = \alpha \hat{S} \hat{Y}^{(t)} + (1 - \alpha) Y
    \label{equ:iterative}
\end{equation}
until convergence, that refines initial node predictions $Y \in \real^{N \times C}$ across $C$ classes based on the graph structure\footnote{A common use of LP is to propagate from labeled to unlabeled nodes; instead we refine predictions for all  nodes which are already labeled.}.
Hyper-parameter $\alpha \in (0,1)$ controls the amount of propagation and refinement. It is possible to show~\cite{zhou03lp} that this iterative solution is equivalent to solving $C$ $N$-variable linear systems
\begin{equation}
    \cL(S) \hat{Y} = Y,
    \label{equ:lp_system}
\end{equation}
where $\cL(S) = I - \alpha D^{-\frac{1}{2}} S D^{-\frac{1}{2}} = I - \alpha \hat{S}$ is the graph Laplacian. The closed-form solution of these linear systems
\begin{equation}
    \hat{Y} = \cL(S)^{-1}Y = \Linv(S) Y,
    \label{equ:lp_closed}
\end{equation}
is not practical for large graphs as $\Linv(S)$ is a dense $\real^{N \times N}$ matrix. However, given that $L$ is positive-definite, it is usual~\cite{iscen17efficient,iscen19lpsemi} to solve \equ{lp_system} using the conjugate-gradient method which is known to be faster than running the iterative solution~\cite{iscen17efficient}. 
Matrix $\Linv$ provides geodesic similarities between all pairs of nodes, as these are captured based on the graph structure. Additionally, label propagation corresponds to minimizing the quadratic criterion
\begin{equation}
\small
    Q(\hat{Y}) = (1 - \alpha) \sum_{i=1}^{N} \left \Vert \hat{Y}_i - Y_i \right \Vert ^2 \\
    + \alpha \sum_{i,j=1}^{N} S_{ij} \left \Vert \frac{\hat{Y}_i}{\sqrt{D_{ii}}} - \frac{\hat{Y}_j}{\sqrt{D_{jj}}} \right \Vert ^2.
    \label{equ:lp_criterion}
\end{equation}
The criterion \equ{lp_criterion} consists of two parts: the first one keeps node predictions close to the initial ones, and the second one pushes neighboring nodes in the graph to have the same predictions.

\noindent\textbf{Exploiting a Vision Model (VM).}
VLMs are trained by minimizing cross-modal similarities between corresponding visual and textual features. This training approach results in suboptimal performance on dense prediction tasks, such as open vocabulary semantic segmentation~\cite{zhou2022extract}, for two main reasons. First, VLMs are typically trained using a loss on \textit{global} image representations, which means they do not explicitly optimize patch-level features. Second, VLM training focuses solely on cross-modal similarities and does not account for relationships across patches, making the resulting representations less effective for capturing the inter-patch relationships essential for semantic segmentation.

For these reasons, recent works~\cite{wysoczanska2024clip,lan2024proxyclip,kang2024defense} further utilize a separate \textit{Vision Model (VM)} to provide improved patch-level representations. 
Let $Z_\text{vm} = h(X)$ denote patch-level features from such a model, where $h: \cI \rightarrow \real^{d' \times N_P}$ is an encoder with a ViT architecture.
The usual choice for the vision model is DINO~\cite{caron21dino}, an encoder trained with self-supervised learning that exhibits strong performance for dense prediction tasks~\cite{caron21dino,simeoni21lost,simeoni23found}.

The training-free variant of the CLIP-DINOiser~\cite{wysoczanska2024clip} 
improves VLM predictions $Y_{\text{vlm}}$ by refining VLM image features $Z_{\text{vlm}}$ using patch relations from VM. They use image features $Z_{\text{vm}}$ and compute an affinity matrix $A = Z_{\text{vm}}^\top Z_{\text{vm}}$. This affinity is then used to refine the VLM predictions as
\begin{equation}
    Y_{\text{DINOiser}} = A Z_\text{vlm}^{\top} F = A Y_{\text{vlm}} .
    \label{equ:dinoiser}
\end{equation}
This process propagates the patch features, or, equivalently, due to linearity, the VLM predictions, over the affinity matrix\footnote{Their implementation $\l2$-normalizes features after propagation, \ie $\l2(AX)F^{\top}$, which breaks the equivalence but performs slightly better.}.
This is a smoothening operation, which we also perform in our method but in a more principled way.

\noindent\textbf{Class prediction using a sliding window approach.}
Current approaches~\cite{wysoczanska2024clip,kang2024defense,lan2024proxyclip,wang2024sclip} perform feature extraction and prediction in a sliding-window fashion to handle test images of arbitrary resolution. Let $\mathcal{W}$ denote a set of $K$ sliding windows. Prediction typically proceeds in 3 steps:
\begin{enumerate}
\item Extract features $Z_\text{vlm}^{(w_i)}$ and obtain predictions $Y_\text{vlm}^{(w_i)} = {Z_\text{vlm}^{(w_i)}}^\top F$ independently for each window $w_i \in \mathcal{W}$.
\item Upsample patch-level predictions to the original window resolution using function $\texttt{up}(Y_\text{vlm}^{(w_i)})$.
\item Obtain per-pixel predictions by averaging predictions over all sliding windows that contain each pixel and combine them into an $H \times W$ class confidence map $Y_\text{vlm}^{(img)} = \texttt{combine}(\{\texttt{up}(Y_\text{vlm}^{(w_1)}), \ldots, \texttt{up}(Y_\text{vlm}^{(w_K)})\})$ .
\end{enumerate}

The sliding window approach presented above is a result of the poor generalization of ViTs to resolutions and aspect ratios different from the ones used during training. 
All existing approaches couple feature extraction with class prediction and perform both \textit{jointly}, and \textit{independently in each window} before averaging predictions across windows. In the following sections, we decouple the two for our approach and perform predictions by looking across windows.

\begin{figure*}
    \centering
    \vspace{-10pt}
    \includegraphics[width=\textwidth]{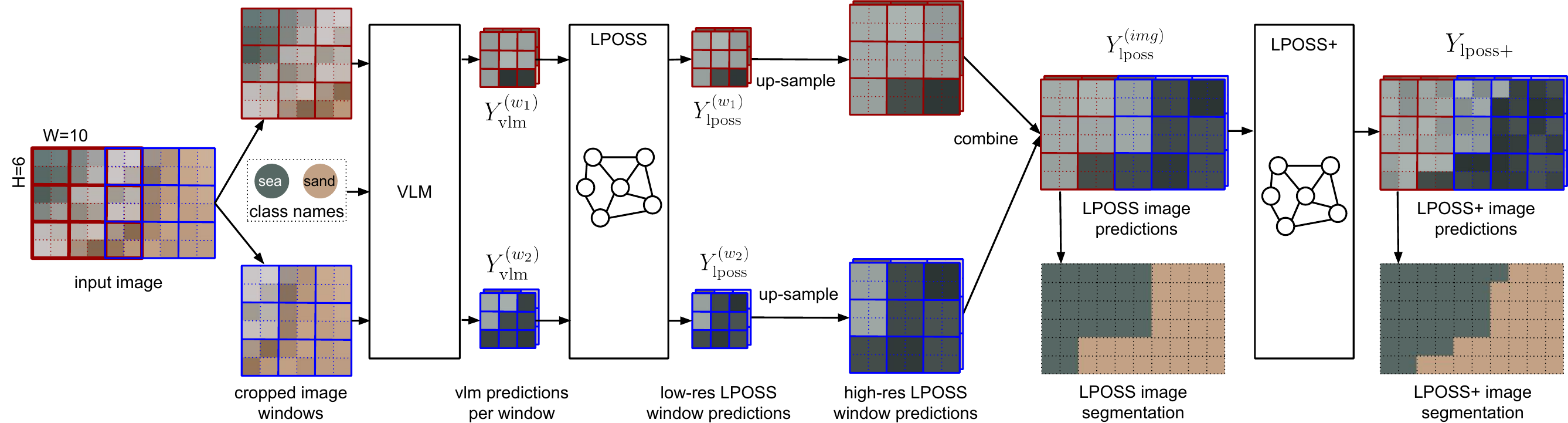}
    \caption{\textbf{Overview of \ours and \oursplus. }     
    Processing steps:
    (1) input image window  cropping 
    (2) feature extraction with the VLM and patch-level predictions per window
    (3) patch-level predictions jointly across all windows with \ours 
    (4) upsampling of patch-level predictions
    (5) combining window predictions into a single pixel-level image prediction
    (6) pixel-level prediction with \oursplus.
    The graphs of \ours and \oursplus are generated based on patch features extracted with the VM and pixel color values, respectively. 
    \label{fig:lp_illustrate}
    \vspace{-15pt}}    
\end{figure*}

\begin{figure}
    \centering
    \begin{tabular}{@{\xssp}c@{\msp}c}
    &79.0/67.8\\
    \includegraphics[width=.15\textwidth]{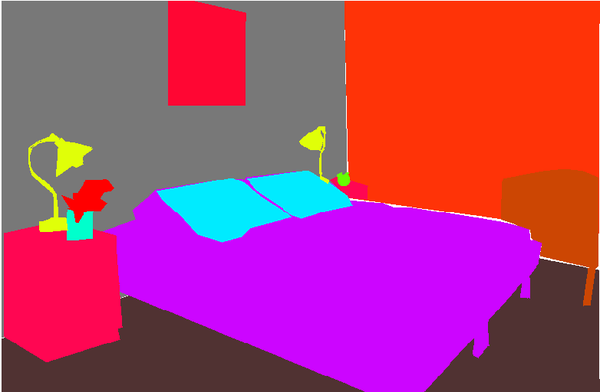}&
    \includegraphics[width=.15\textwidth]{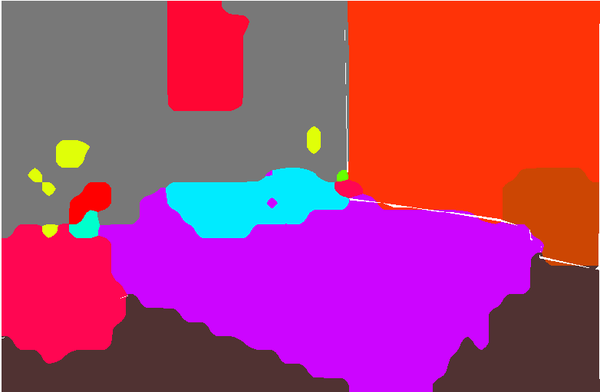}\\
    &32.8/6.9\\
    \includegraphics[width=.15\textwidth]{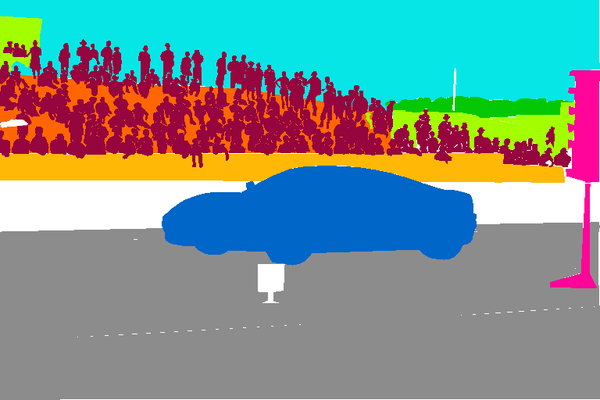}&
    \includegraphics[width=.15\textwidth]{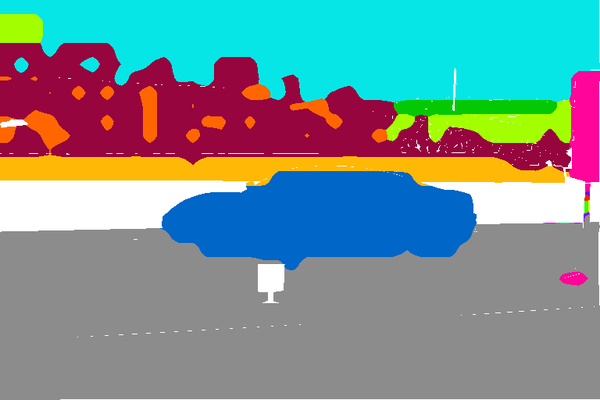}\\ 
    ground-truth & ground-truth at patch-level res.\\
    \end{tabular}
    \caption{\textbf{Visualization of artifacts caused by patch-level predictions.} Ground-truth at patch-level resolution produced by downsampling the ground-truth by the patch size $P=16$ and then upsampling to the original size. Performance reported via mIoU / Boundary IoU. 
    \label{fig:impact_of_patches}
    \vspace{-15pt}}
\end{figure}

\subsection{\ours: LP over patches}
\label{sec:lp_patches}
We observe that~\equ{dinoiser} and \equ{lp_closed} can both be seen as different forms of label refinement. In~\equ{dinoiser}, the VLM-based labels $Y_\text{vlm}$ are refined using matrix $A$ of Euclidean similarities across patches, while in~\equ{lp_closed}, graph labels $Y$ are refined via the geodesic similarities $\Linv$, computed over the graph. 
This motivates us to utilize label propagation to refine the VLM predictions. Not only is \equ{lp_closed} optimizing \equ{lp_criterion}, which is a principled and intuitive objective function, but we also expect geodesic similarities to capture more complex relations and contextual information, as they are computed by considering the entire set of patch features together.

We assume a patch $\pp_i = \left(\vz_{\text{vm}_{i}}, \vp_i\right)$ is represented as a tuple by its features $\vz_{\text{vm}_{i}}$ from the VM, \ie the $i$-th column of $Z_\text{vm}$, and the location of its center $\vp_i$ in the image. We then construct a graph with nodes $\{\pp_1,  \ldots, \pp_{N_P}\}$ and adjacency matrix $S_{\mathtt{P}} = S_a \odot S_p$, where $\odot$ denotes element-wise multiplication. $S_a$ is an appearance-based adjacency matrix and $s_{a_{ij}} = (\vz_{\text{vm}_i}^{\top} \vz_{\text{vm}_j})^{\gamma}$ if $\vz_{\text{vm}_j}$ is in the k-nearest neighbors of $\vz_{\text{vm}_i}$, and 0 otherwise. $S_p$ is a position-based adjacency matrix that depends on the spatial distance between patches with $s_{p_{ij}} = \exp\left(-\frac{||\vp_i - \vp_j||^2}{\sigma}\right)$. Having such a graph, we apply label propagation by 
\begin{equation}
Y_\text{lposs} = \Linv(S_{\mathtt{P}}) Y_\text{vlm}
    \label{equ:lposs}
\end{equation}
to refine the VLM predictions. This process effectively refines the predictions such that neighboring patches (patches with high appearance similarity or spatial proximity) have similar predictions, while keeping the final prediction not far from the VLM predictions.

\subsection{\ours across all windows}
We propose to decouple the feature extraction process from class prediction. First, we perform feature extraction on a per-window basis to be consistent with the encoder’s training resolution. Unlike existing methods, however, we then carry out predictions \textit{across windows}, \ie, jointly for the entire image. This approach enables us to explicitly account for interactions across all patches.
 
Sliding-window extraction provides us with baseline VLM predictions $Y_\text{vlm}^{(w_i)}$ for window $w_i$ with patches $\pp^{(w_i)}_j, j=1,\ldots, N_P$, which we merge in
$Y_\text{vlm}^{(\mathcal{W})} = [{Y_\text{vlm}^{(w_1)}}^\top, \ldots, {Y_\text{vlm}^{(w_K)}}^\top]^\top$, across all windows.
We construct affinity matrix $S^{(\mathcal{W})} \in \real^{KN_P \times KN_P}$ over nodes
 $\{\pp^{(w_1)}_1,  \ldots, \pp^{(w_1)}_{N_P}, \ldots, \pp^{(w_K)}_1,  \ldots, \pp^{(w_K)}_{N_P}\}$
 containing patches from all sliding windows, 
 and perform LP jointly for the whole image by 
 \begin{equation}
Y_\text{lposs}^{(\mathcal{W})} = \Linv(S^\mathcal{W}) Y_\text{vlm}^{(\mathcal{W})}.
\label{equ:lposs_windows}
 \end{equation}
We decompose the output prediction into parts associated with each window by
$Y_\text{lposs}^{(\mathcal{W})} = [{Y_\text{lposs}^{(w_1)}}^\top, \ldots, {Y_\text{lposs}^{(w_K)}}^\top]^\top$,
and, similar to other approaches, upsample and combine by
\begin{equation}
    Y_\text{lposs}^{(img)} = \texttt{combine}(\{\texttt{up}(Y_\text{lposs}^{(w_1)}), \ldots, \texttt{up}(Y_\text{lposs}^{(w_K)})\}),
\end{equation}
with  $Y_\text{lposs}^{(img)} \in \real^{H \times W \times C}$.
We show a graphical representation of this process in Figure~\ref{fig:lp_illustrate}.

\subsection{\oursplus: LP over pixels}
Similar to related approaches, \ours
performs predictions at patch resolution, which are subsequently upsampled to the original image resolution. However, we observe that this process results in block-like artifacts in the predictions, as shown in Figure~\ref{fig:impact_of_patches}.

To assess the impact of these artifacts on performance, we conducted the following experiment on the ground-truth maps: we downsampled the ground-truth map to patch-level resolution by a factor corresponding to the patch size (e.g., $P=16$), then upsampled it back to the original image resolution. We treat the result as a prediction and evaluate performance using mIoU and Boundary IoU~\cite{cheng2021boundary} relative to the ground truth. The average performance across eight datasets was approximately $85\%$ for mIoU and $70\%$ for Boundary IoU, as shown in Table~\ref{tab:sota_results} and Table~\ref{tab:boundary_iou}, indicating that patch-level predictions inherently limit performance.
\looseness=-1 To address this, we propose to apply label propagation 
at the pixel level 
to further refine \ours predictions $Y_\text{lposs}^{(img)}$.

We assume a pixel $\tilde{\pp}_i = \left(\tilde{\vz}_i, \tilde{\vp}_i \right)$ to be represented as a tuple by its location $\tilde{\vp}_i$ within the image and its features $\tilde{\vz}_i$, which may be based on the RGB values or any encoder. Similar to the patch-level case, we construct a graph with pixels as nodes and adjacency matrix $S_{\tilde{\mathtt{P}}} = S_{\tilde{a}} \odot S_{\tilde{p}}$.
The position-based affinity $S_{\tilde{p}}$ is binary, with non-zero elements within the $r \times r$ neighborhood of a pixel, and the appearance-based affinity $S_{\tilde{a}}$ has elements equal to 
$s_{\tilde{a}_{ij}} = \exp\left(-\frac{||\tilde{\vz}_i - \tilde{\vz}_j||^2}{\tau}\right)$.
 
We apply label propagation on top of the pixel-level graph by
\begin{equation}
Y_\text{lposs+}  = \Linv( S_{\tilde{\mathtt{P}}} ) Y_\text{lposs}^{(img)},
\label{equ:lpossplus}
\end{equation}
with $Y_\text{lposs+}^{(img)} \in \real^{H \times W \times C}$, which, after $\argmax$, becomes the final segmentation mask.
This process refines the predictions such that neighboring pixels have similar predictions, while keeping the final prediction not far from the \ours predictions. We call this approach \textbf{\oursplus} and show its visualization in Figure~\ref{fig:lp_illustrate}.

\begin{table*}[!h]
    \centering
    {
\footnotesize
\newcommand{\cmark}{\ding{51}}%
\newcommand{\xmark}{\ding{55}}
\begin{tabular}{lccccccccc@{\hspace{25pt}}c}
\toprule
Method & VM & VOC & Object & Context & C59 & Stuff & VOC20 & ADE20k & City & Avg \\
\midrule
    Oracle (patch-level res.) & - & 91.5 &	83.8 &	86.6 &	86.8 &	86.0 &	92.8 &	80.6 &	73.6 &	85.2 \\
\midrule
    MaskCLIP\textsuperscript{*}~\cite{zhou2022extract} & \xmark & 32.9 & 16.3 & 22.9 & 25.5 & 17.5 & 61.8 & 14.2 & 25.0 & 27.0 \\
    GEM~\cite{bousselham24gem} & \xmark & 46.8 & -  & 34.5 & - & - & - & 17.1 & - & - \\
    SCLIP~\cite{wang2024sclip} & \xmark & 59.1 & 30.5 & 30.4 & 34.2 & 22.4 & 80.4 & 16.1 & 32.2 & 38.2 \\
    ClearCLIP~\cite{lan2024clearclip} & \xmark & 51.8 & 33.0 & 32.6 & 35.9 & 23.9 & \underline{80.9} & 16.7 & 30.0 & 38.1 \\

    CLIP-DIY\textsuperscript{\dag}~\cite{wysoczanska2024diy} & \cmark & 59.9 & 31.0 & 19.7 & 19.8 & 13.3 & 79.7 & 9.9 & 11.6 & 30.6 \\
    CLIP-DINOiser\textsuperscript{*}~\cite{wysoczanska2024clip} & \cmark & \underline{62.2} & \underline{34.7} & 32.5 & 36.0 & 24.6 & 80.8 & 20.1 & 31.1 & 40.2 \\
    ProxyCLIP\textsuperscript{*}~\cite{lan2024proxyclip} & \cmark & 59.3 & \textbf{36.3} & 34.4 & \underline{38.0} & 25.7 & 79.7 & 19.4 & 36.0 & 41.1 \\
    LaVG\textsuperscript{*}~\cite{kang2024defense} & \cmark & 61.8 & 33.3 & 31.5 & 34.6 & 22.8 & \textbf{81.9} & 14.8 & 25.0 & 38.2 \\\midrule
    \ours & \cmark & 61.1 & 33.4 & \underline{34.6} & 37.8 & \underline{25.9} & 78.8 & \underline{21.8} & \underline{37.3} & \underline{41.3} \\
    \oursplus & \cmark & \textbf{62.4} & 34.3 & \textbf{35.4} & \textbf{38.6} & \textbf{26.5} & 79.3 & \textbf{22.3} & \textbf{37.9} & \textbf{42.1}\\
\bottomrule
\end{tabular}
}

    \caption{\textbf{Performance comparison in terms of mIoU on 8 datasets} using ViT-B/16 backbone for both VLM and VM. \textsuperscript{*} denotes the methods for which we reproduce the performance. \textsuperscript{\dag} denotes numbers taken from Wysoczanska \etal~\cite{wysoczanska2024clip}.}
    \label{tab:sota_results}
\end{table*}

\begin{table*}[!h]
    \centering
    {
\footnotesize
\newcommand{\cmark}{\ding{51}}%
\newcommand{\xmark}{\ding{55}}
\begin{tabular}{lccccccccc@{\hspace{25pt}}c}
\toprule
Method & VM & VOC & Object & Context & C59 & Stuff & VOC20 & ADE20k & City & Avg \\
\midrule
    Oracle (patch-level res.) & - & 80.1 &	61.3 &	62.6 &	68.0 &	69.0 &	83.5 &	65.0 &	66.4 &	69.5 \\
\midrule
    MaskCLIP~\cite{zhou2022extract} & \xmark & 17.2 &	7.0 &	9.2 &	12.0 &	9.3 &	38.6 &	8.2 &	17.2 &	14.8 \\
    CLIP-DINOiser~\cite{wysoczanska2024clip} & \cmark & 47.4 &	17.5 &	15.7 &	22.4 &	15.6 &	71.1 &	12.3 &	22.1 &	28.0 \\
    ProxyCLIP~\cite{lan2024proxyclip} & \cmark & 45.0 &	\underline{19.7} &	17.4 &	23.0 &	15.9 &	69.4 &	12.9 &	\underline{28.4} &	29.0 \\
    LaVG~\cite{kang2024defense} & \cmark & \underline{49.5} &	19.1 &	18.3 &	23.6 &	15.4 &	\textbf{74.3} &	9.2 &	15.7 &	28.1 \\
    \midrule
    \ours & \cmark & 48.5 &	18.4 &	\underline{19.1} &	\underline{24.8} &	\underline{17.9} &	70.1 &	\underline{14.9} &	\underline{28.4} &	\underline{30.3} \\
    \oursplus & \cmark & \textbf{51.2} &	\textbf{20.4} &	\textbf{21.6} &	\textbf{27.1} &	\textbf{19.2} &	\underline{71.7} &	\textbf{16.0} &	\textbf{29.5} &	\textbf{32.1}\\
\bottomrule
\end{tabular}
}

    \caption{\textbf{Performance comparison in terms of Boundary IoU on 8 datasets} using ViT-B/16 backbone both for VLM and VM.
    }
    \label{tab:boundary_iou}
\end{table*}

\section{Experiments}

\subsection{Datasets and evaluation metrics}

We evaluate our method on eight datasets: PASCAL VOC~\cite{pascal-voc-2012} (VOC), COCO Object~\cite{caesar18coco} (Object), PASCAL Context~\cite{mottaghi14context} (Context), PASCAL Context59~\cite{mottaghi14context} (C59), COCO Stuff~\cite{caesar18coco} (Stuff), PASCAL VOC20~\cite{pascal-voc-2012} (VOC20), ADE20k~\cite{zhou17scene,zhou19semantic}, and Cityscapes~\cite{cordts16cityscapes} (City), that are commonly used~\cite{wysoczanska2024clip,wang2024sclip,lan2024proxyclip,kang2024defense} to evaluate open-vocabulary semantic segmentation. Following standard practice, we report mean Intersection over Union (mIoU) metric and additionally Boundary IoU~\cite{cheng2021boundary} that measures performance only near the ground-truth mask boundaries.

\subsection{Experimental setup}

\noindent\textbf{Backbones and textual prompts.}
We report the results using ViT-B/16 backbones and use DINO~\cite{caron21dino} as a visual model $h$ and OpenCLIP~\cite{cherti23openclip} as a VLM $(f, g)$. We follow the setup of CLIP-DINOiser~\cite{wysoczanska2024clip} and use the output of DINO's last layer's \emph{value embedding} as the visual model features $Z_{\text{vm}}$, while we use the output of MaskCLIP~\cite{zhou2022extract} as the VLM features $Z_{\text{vlm}}$ and $F$. We use ImageNet prompt templates from CLIP~\cite{radford21clip} as templates for the VLM, following standard practice~\cite{wysoczanska2024clip,kang2024defense,lan2024proxyclip,zhou2022extract}. For datasets that have a class \texttt{background}, we use text expansion for it and expand \texttt{background} into, \eg \texttt{sky}, \texttt{wall}, \etc, following SCLIP~\cite{wang2024sclip}, ProxyCLIP~\cite{lan2024proxyclip}, and LaVG~\cite{kang2024defense}.

\noindent\textbf{Compared methods}
\looseness=-1 We compare our method with methods that use only a VLM, by making small changes during inference to the general ViT architecture, for open-vocabulary semantic segmentation: \textbf{MaskCLIP}~\cite{zhou2022extract}, \textbf{GEM}~\cite{bousselham24gem}, \textbf{SCLIP}~\cite{wang2024sclip}, and \textbf{ClearCLIP}~\cite{lan2024clearclip}. Additionally, we compare with methods that utilize  vision models to aid VLMs: \textbf{CLIP-DIY}~\cite{wysoczanska2024diy}, \textbf{CLIP-DINOiser}~\cite{wysoczanska2024clip}\footnote{ 
We use their trained model in our evaluation because it is publicly available and performs similarly to the training-free variant.
}, \textbf{ProxyCLIP}~\cite{lan2024proxyclip}, and \textbf{LaVG}~\cite{kang2024defense}.

\noindent\textbf{Implementation details}
We reproduce results for CLIP-DINOiser\footnote{\url{https://github.com/wysoczanska/clip_dinoiser}}, ProxyCLIP\footnote{\url{https://github.com/mc-lan/ProxyCLIP}}, and LaVG\footnote{\url{https://github.com/dahyun-kang/lavg}} following their official implementations, while we reproduce MaskCLIP using the implementation provided by CLIP-DINOiser. We report the performance of CLIP-DIY~\cite{wysoczanska2024diy} as reported in CLIP-DINOiser~\cite{wysoczanska2024clip}, while for GEM~\cite{bousselham24gem}, SCLIP~\cite{wang2024sclip}, and ClearCLIP~\cite{lan2024clearclip} we report the numbers provided by the original papers. We implement our method using MMSegmentation~\cite{mmseg20mmseg} using the sliding-window approach with window size $224 \times 224$, window stride $112 \times 112$, and resizing the input image to have the shorter side of $448$ for all datasets. We also fix the values of hyper-parameters $\alpha$, $k$, $\gamma$, $\sigma$, $r$, and $\tau$ for \ours and \oursplus to $0.95$, $400$, $3.0$, $100$, $13$, and $0.01$, respectively, across all datasets. For pixel features of \oursplus, we use image pixels converted to Lab color space. We discuss the design choices and show the impact of different hyper-parameters of \ours and \oursplus in Section~\ref{sec:desing_lposs} of the supplementary.

{
\setlength{\tabcolsep}{-1.5pt}
\begin{figure*}
\footnotesize
   \begin{center}
   \begin{tabular}{c@{\ssp}c@{\ssp}c@{\ssp}c@{\ssp}c@{\ssp}c@{\ssp}c}
      Image & \textbf{GT} & \textbf{MaskCLIP} & \textbf{CLIP-DINOiser} & \textbf{ProxyCLIP} & \textbf{LaVG} & \textbf{\oursplus} \\ 
      \includegraphics[width=0.14\textwidth]{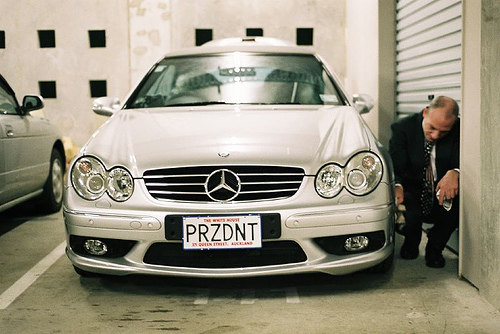}
      & 
      \includegraphics[width=0.14\textwidth]{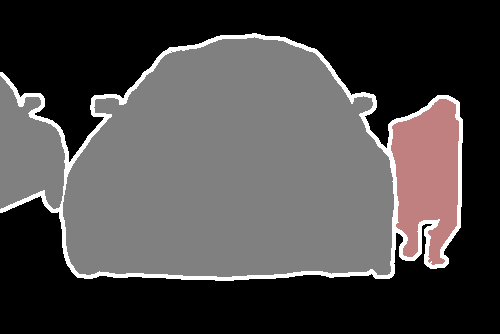}
      & 
      \includegraphics[width=0.14\textwidth]{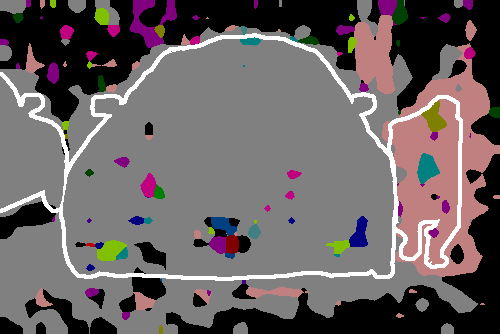}
      & 
      \includegraphics[width=0.14\textwidth]{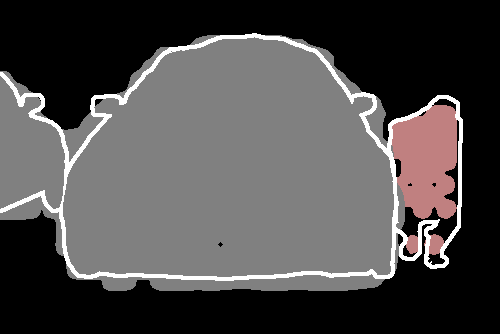}
      & 
      \includegraphics[width=0.14\textwidth]{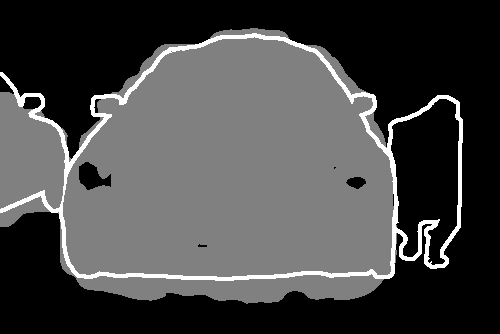}
      & 
      \includegraphics[width=0.14\textwidth]{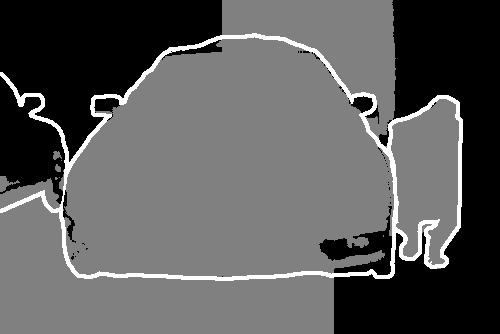}
      & 
      \includegraphics[width=0.14\textwidth]{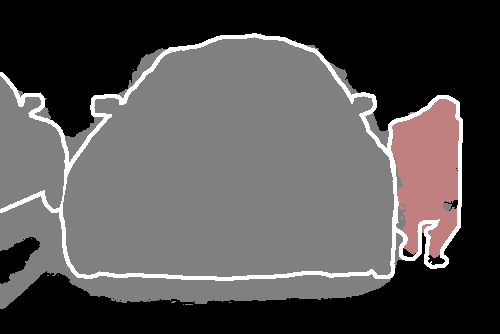}
      \\
       & & 9.7/4.0 & 85.6/42.5 & 56.2/25.8 & 36.3/19.7 & 87.5/49.0 \\ 
      \includegraphics[width=0.14\textwidth]{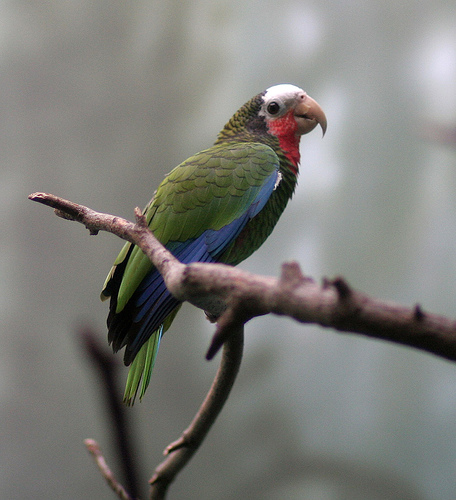}
      & 
      \includegraphics[width=0.14\textwidth]{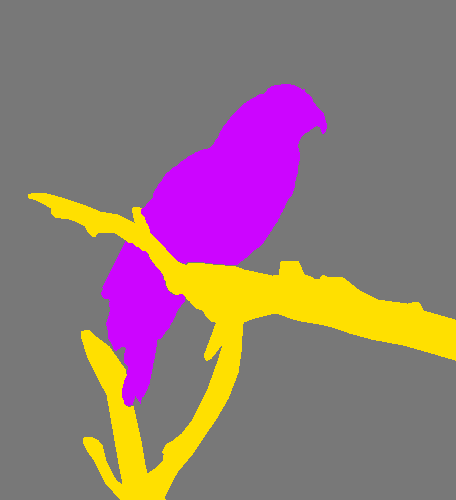}
      & 
      \includegraphics[width=0.14\textwidth]{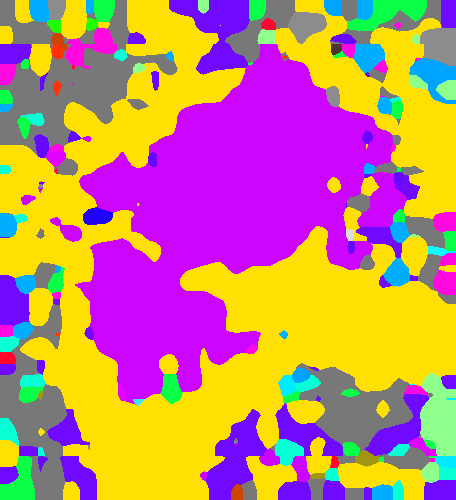}
      & 
      \includegraphics[width=0.14\textwidth]{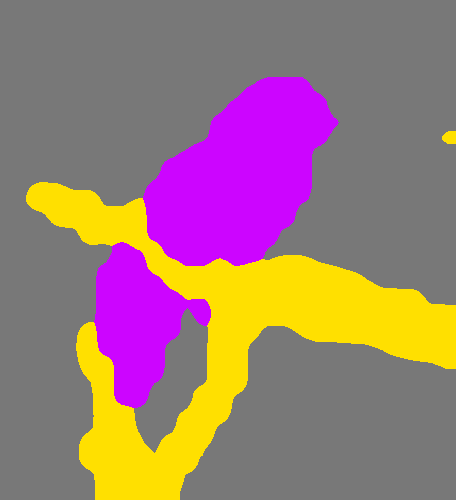}
      & 
      \includegraphics[width=0.14\textwidth]{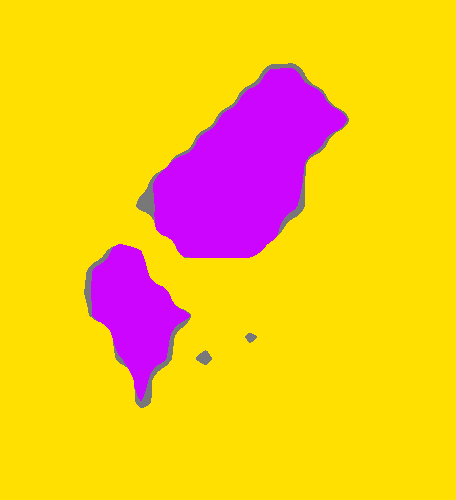}
      & 
      \includegraphics[width=0.14\textwidth]{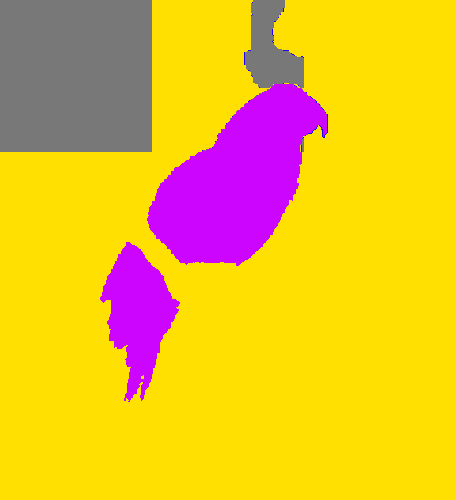}
      & 
      \includegraphics[width=0.14\textwidth]{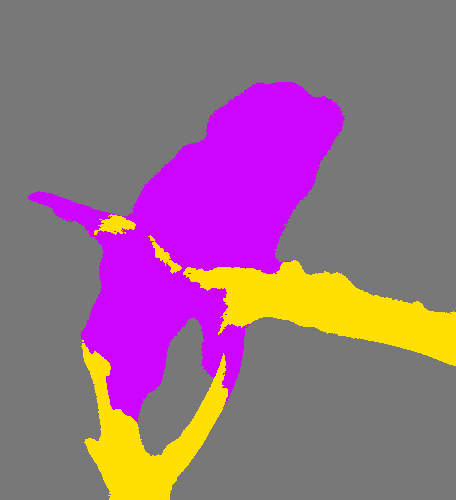}
      \\
       & & 3.2/1.0 & 75.3/41.7 & 30.4/18.5 & 30.9/25.2 & 72.7/47.7 \\ 
      
      \includegraphics[width=0.14\textwidth]{fig/visualization_masks/coco_stuff_000000186422/000000186422_img.png}
      & 
      \includegraphics[width=0.14\textwidth]{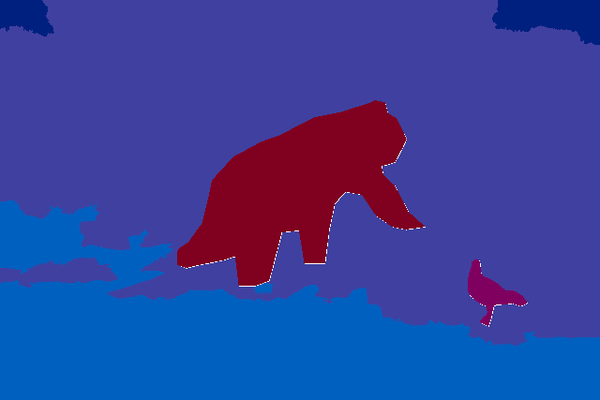}
      & 
      \includegraphics[width=0.14\textwidth]{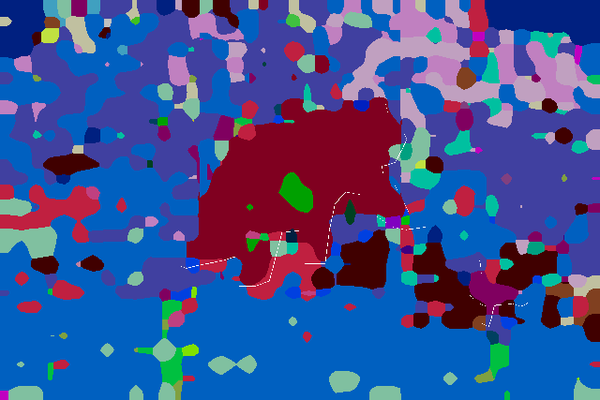}
      & 
      \includegraphics[width=0.14\textwidth]{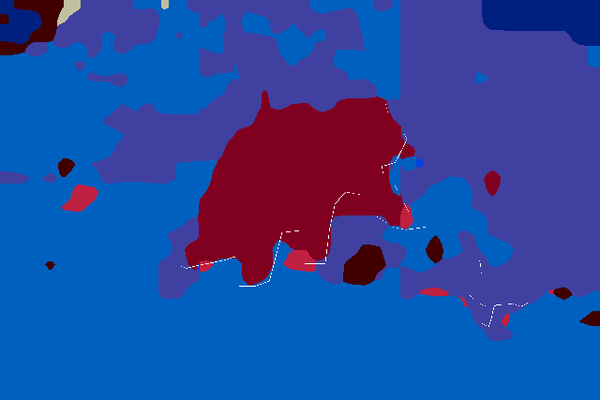}
      & 
      \includegraphics[width=0.14\textwidth]{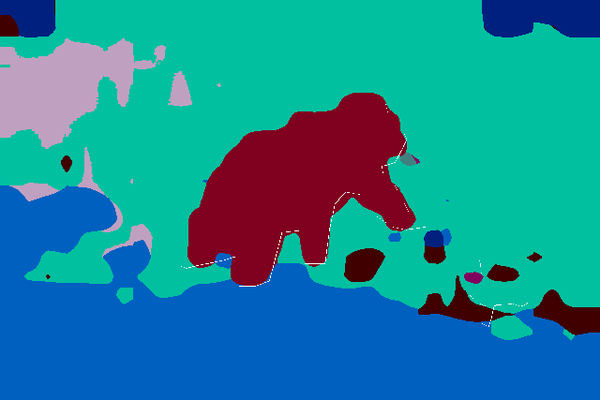}
      & 
      \includegraphics[width=0.14\textwidth]{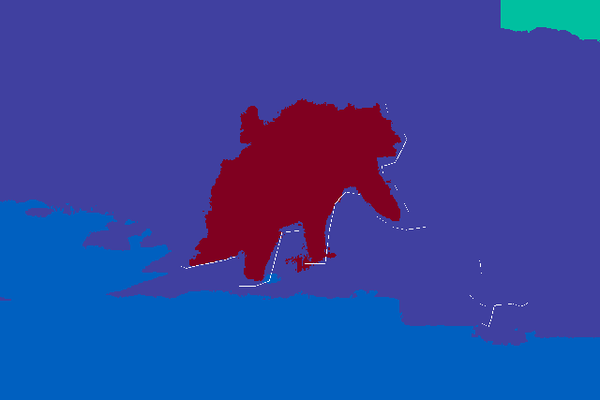}
      & 
      \includegraphics[width=0.14\textwidth]{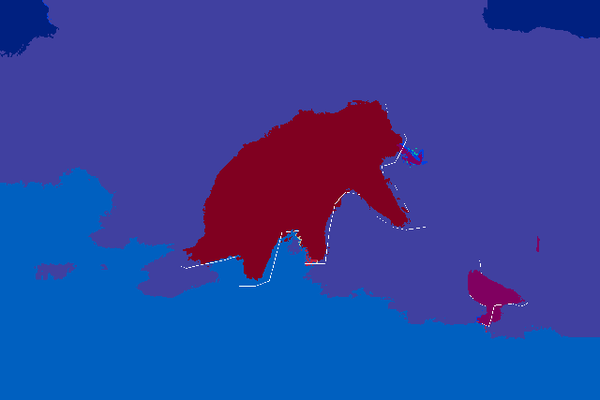}
      \\
       & & 5.9/3.9 & 28.1/17.1 & 25.2/19.0 & 43.6/31.2 & 45.3/36.2 \\ 
      \includegraphics[width=0.14\textwidth]{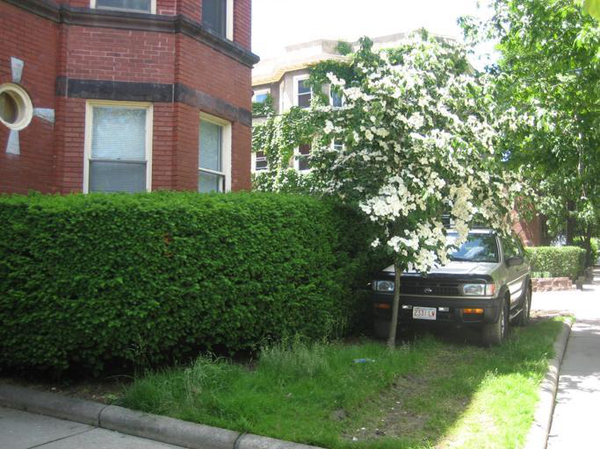}
      & 
      \includegraphics[width=0.14\textwidth]{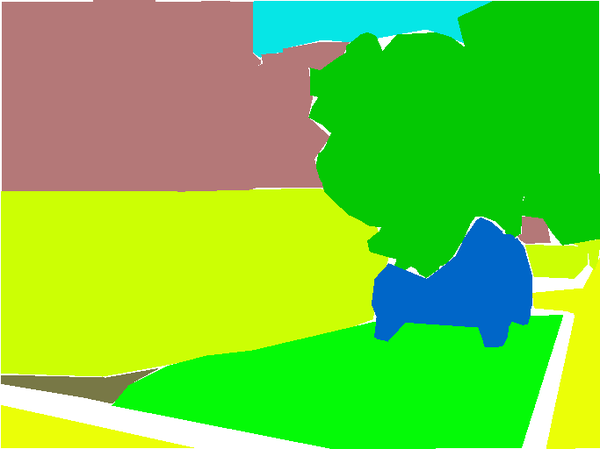}
      & 
      \includegraphics[width=0.14\textwidth]{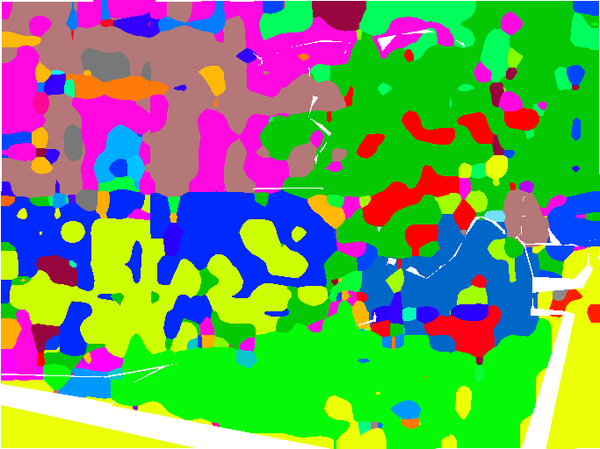}
      & 
      \includegraphics[width=0.14\textwidth]{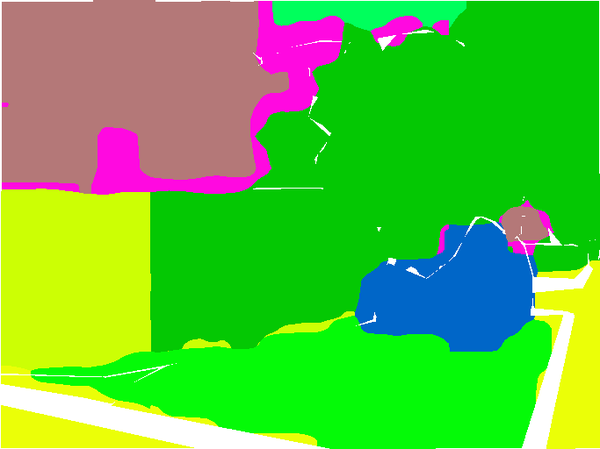}
      & 
      \includegraphics[width=0.14\textwidth]{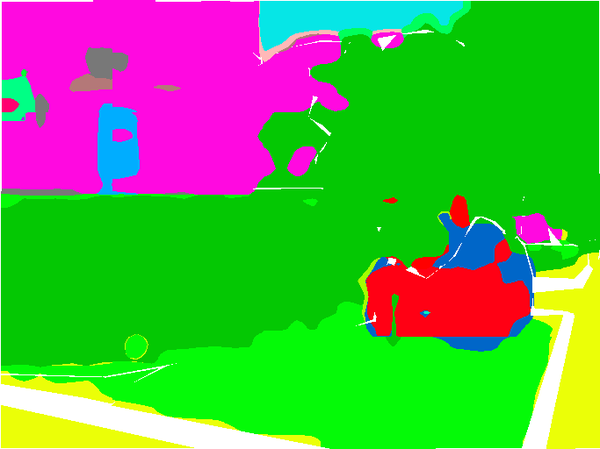}
      & 
      \includegraphics[width=0.14\textwidth]{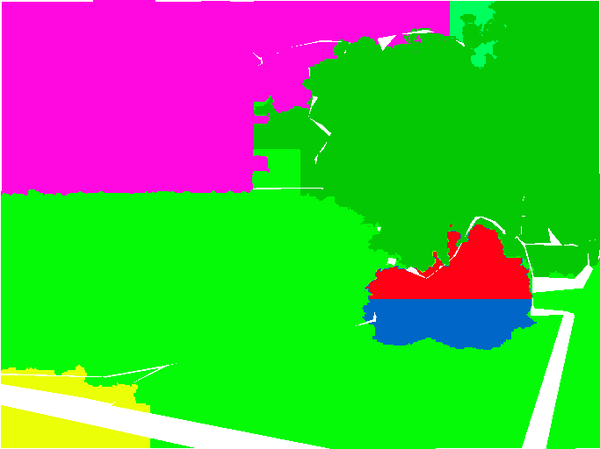}
      & 
      \includegraphics[width=0.14\textwidth]{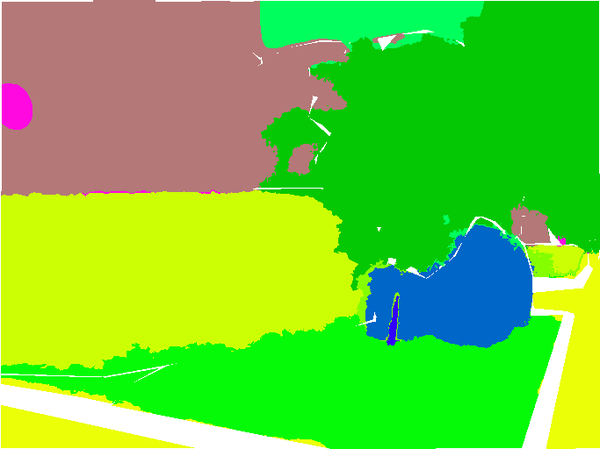}
      \\
       & & 5.7/2.2 & 41.2/20.2 & 13.9/8.4 & 15.3/8.2 & 42.7/23.1 \\ 
      \includegraphics[width=0.14\textwidth]{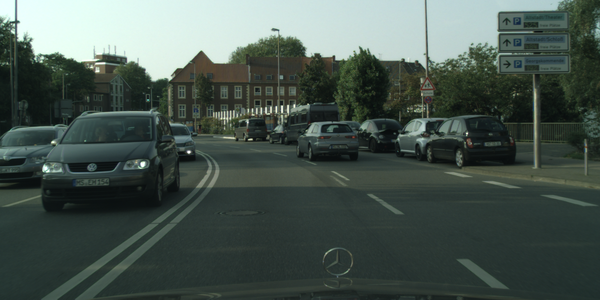}
      & 
      \includegraphics[width=0.14\textwidth]{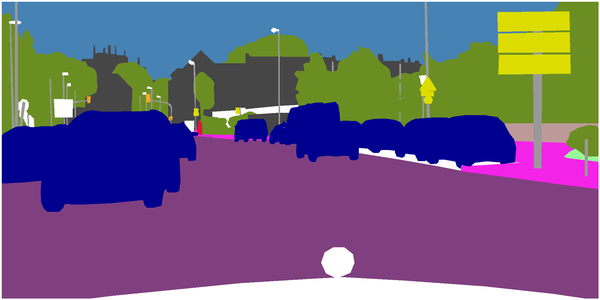}
      & 
      \includegraphics[width=0.14\textwidth]{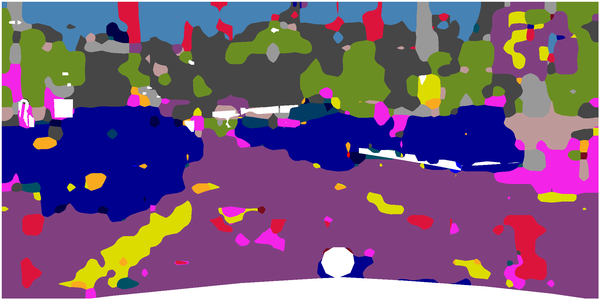}
      & 
      \includegraphics[width=0.14\textwidth]{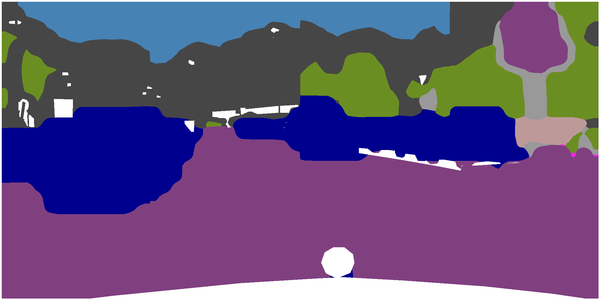}
      & 
      \includegraphics[width=0.14\textwidth]{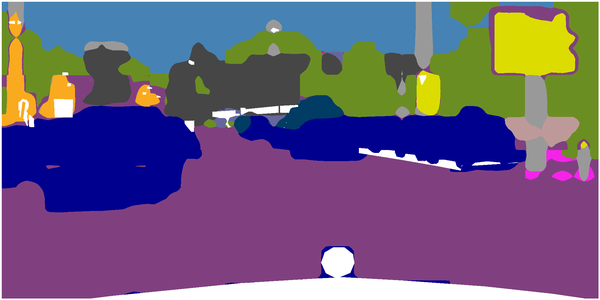}
      & 
      \includegraphics[width=0.14\textwidth]{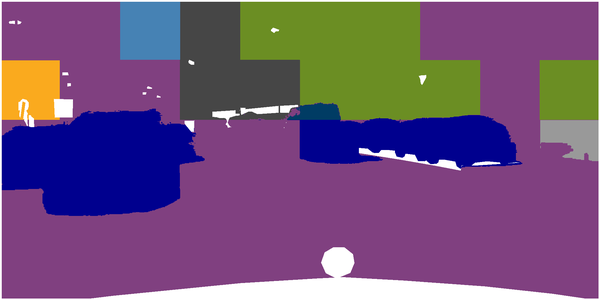}
      & 
      \includegraphics[width=0.14\textwidth]{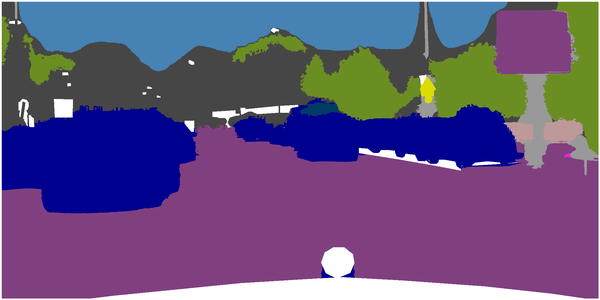} \\
      & & 19.4/13.2 & 30.6/19.5 & 38.8/25.5 & 18.1/10.1 & 34.5/25.0 \\ 
   \end{tabular}
\end{center}
\caption{\textbf{Qualitative comparison of open-vocabulary semantic segmentation.} A comparison of \oursplus with the best performing methods: MaskCLIP~\cite{zhou2022extract}, CLIP-DINOiser~\cite{wysoczanska2024clip}, ProxyCLIP~\cite{lan2024proxyclip}, and LaVG~\cite{kang2024defense}. On top of each segmentation map we show its mIoU/Boundary IoU. Pixels shown in white are pixels that do not have a class in the ground-truth.
\vspace{-10pt}
}
    \label{fig:segmentation_visual}
\end{figure*}
}

\subsection{Results}
\label{sec:results}
We present the quantitative results in Table~\ref{tab:sota_results} given by the conventionally used mIoU metric. \ours achieves state-of-the-art performance on average, with competitive performance across most datasets with only a few notable exceptions. One such exception happens on the VOC20 dataset where \ours has low performance compared to others. We observe this is the case because of the choice of the window size, something that we discuss in other experiments. \oursplus further improves the results of \ours by $0.8\%$ on average and obtains state-of-the-art performance on all datasets besides VOC20 and Object. 
ProxyCLIP is the only method that significantly outperforms \oursplus on the Objects dataset, something that we attribute to the fact that it employs dataset-specific thresholds to filter the background class in this case, as well as for the VOC dataset.

\begin{table*}[t]
    \centering
    \footnotesize
\begin{tabular}{lccccccccc}
\toprule
Method & VOC & Object & Context & C59 & Stuff & VOC20 & ADE20k & City & Avg \\
\midrule
\footnotesize{\emph{No sliding windows}} \\
    CLIP-DINOiser~\cite{wysoczanska2024clip} & \textbf{60.2} & 	\textbf{32.4} & 	31.3 &	\underline{34.6} &	23.2 &	79.1 & 	18.7 &	28.2 &	\underline{38.5} \\
    ProxyCLIP~\cite{lan2024proxyclip} & 43.1 &	24.2 &	25.9 &	28.8 &	18.7 &	73.1 &	13.6 &	13.6 &	30.1 \\
    LaVG~\cite{kang2024defense} & 24.3 &	16.1 &	12.7 &	13.7 &	10.0 &	32.5 &	8.2 &	6.8 &	15.6 \\
    \ours & 59.2 &	31.2 &	\underline{31.4} &	34.4 &	\underline{23.3} &	\underline{80.7} &	\underline{19.0} &	\underline{28.4} &	38.4 \\
    \oursplus & \underline{59.7} &	\underline{31.7} &	\textbf{32.0} &	\textbf{35.0} & \textbf{23.6} &		\textbf{81.3} &	\textbf{19.3} &	\textbf{28.5} &	\textbf{38.9} \\
\midrule
\footnotesize{\emph{Window size: $448 \times 448$, window stride: $224 \times 224$}} \\
    CLIP-DINOiser~\cite{wysoczanska2024clip} & \textbf{62.2} & \underline{34.7} & 32.5 & 36.0 & 24.6 & 80.8 & 20.1 & 31.1 & 40.2 \\
    ProxyCLIP~\cite{lan2024proxyclip} & 59.9 & \textbf{36.0} & \textbf{34.6} & \textbf{38.2} & \textbf{25.6} & 78.9 & 18.9 & \textbf{33.6} & \underline{40.7} \\
    LaVG~\cite{kang2024defense} & 36.7 & 19.6 & 16.4 & 18.7 & 13.8 & 44.3 & 10.2 & 10.8 & 21.3 \\
    \ours & 61.3 & 32.5 & 32.9 & 36.3 & 25.2 & \underline{82.8} & \underline{20.4} & 31.8 & 40.4 \\
    \oursplus & \underline{62.0} & 33.1 & \underline{33.4} & \underline{36.9} & \underline{25.5} & \textbf{83.0} & \textbf{20.7} & \underline{32.1} & \textbf{40.8} \\
\midrule
\footnotesize{\emph{Window size: $224 \times 224$, window stride: $112 \times 112$}} \\
    CLIP-DINOiser~\cite{wysoczanska2024clip} & 55.4 & 33.0 & 32.2 & 35.7 & 24.8 & 73.1 & 21.1 & 36.3 & 38.9 \\
    ProxyCLIP~\cite{lan2024proxyclip} & 54.9 & 32.9 & 33.2 & 36.4 & 24.3 & 73.0 & 18.6 & 36.0 & 38.7 \\
    LaVG~\cite{kang2024defense} & \underline{62.0} & 32.3 & 31.8 & 34.7 & 22.8 & \textbf{79.8} & 15.4 & 23.1 & 37.7 \\
    \ours & 61.1 & \underline{33.4} & \underline{34.6} & \underline{37.8} & \underline{25.9} & 78.8 & \underline{21.8} & \underline{37.3} & \underline{41.3} \\
    \oursplus & \textbf{62.4} & \textbf{34.3} & \textbf{35.4} & \textbf{38.6} & \textbf{26.5} & \underline{79.3} & \textbf{22.3} & \textbf{37.9} & \textbf{42.1}\\
\midrule
\footnotesize{\emph{Ensemble of two setups above}} \\
    CLIP-DINOiser~\cite{wysoczanska2024clip} & \underline{63.0} & \textbf{35.9} & 34.1 & 37.8 & 26.3 & 79.6 & 22.1 & 35.9 & 41.8 \\
    ProxyCLIP~\cite{lan2024proxyclip} & 59.0 & \underline{35.1} & \underline{35.3} & \underline{38.9} & 26.1 & 78.1 & 19.8 & \underline{36.8} & 41.1 \\
    LaVG~\cite{kang2024defense} & 51.4 & 27.3 & 24.3 & 28.4 & 20.1 & 63.7 & 14.7 & 20.3 & 31.3 \\
    \ours & 62.9 & 34.1 & 35.1 & 38.6 & \underline{26.5} & \underline{82.1} & \underline{22.2} & 36.5 & \underline{42.3} \\
    \oursplus & \textbf{63.9} & 35.0 & \textbf{35.8} & \textbf{39.3} & \textbf{27.0} & \textbf{82.5} & \textbf{22.7} & \textbf{37.0} & \textbf{42.9}\\
\bottomrule
\end{tabular}

    \vspace{-5pt}
    \caption{\textbf{Impact of sliding windows for feature extraction.} Feature extraction is performed with windows of different size/stride or no windows at all. 
    Following the original design choices,
    other methods perform window-based prediction, while \ours and \oursplus perform prediction jointly across all windows.
    Image resize of the smaller side to 448 is used and the reported metric is mIoU.
    \vspace{-10pt}
    }
    \label{tab:window_impace}
\end{table*}

\noindent\textbf{Boundary IoU metric.} Based on Table~\ref{tab:sota_results} we can already see that going beyond patch-level predictions in \ours to pixel-level predictions in \oursplus improves the results. Improvements are even more visible when we look into the Boundary IoU~\cite{cheng2021boundary} metric, presented in Table~\ref{tab:boundary_iou}, which is representative of how well the predicted object boundaries match the ground truth. \oursplus outperforms all other methods and achieves state-of-the-art on almost all datasets. Interestingly, even \ours outperforms competitors in this metric by a large margin. We attribute this to our choice of the window size, as \ours achieves Boundary IoU of $28.9\%$ on average when we apply it with window size $448 \times 448$ and window stride $224 \times 224$, which is comparable to the other approaches.

Additionally, in Table~\ref{tab:sota_results} and Table~\ref{tab:boundary_iou}, we report mIoU and Boundary IoU, respectively, for the oracle-based experiment for patch-level resolution prediction. In this experiment, we downsample the ground-truth map to the patch-level resolution by a factor equal to the patch size and upsample it to the original image resolution, which we treat as predictions. Both mIoU and especially Boundary IoU are significantly impacted by this process, which shows that patch-level predictions are significantly limiting the segmentation performance.

\noindent\textbf{Qualitative results.}
We present the qualitative results of \oursplus and the most important competitors in Figure~\ref{fig:segmentation_visual}. We observe that \oursplus obtains good results across a variety of images coming from different benchmark datasets. 
Qualitatively, our output is closer to that of CLIP-DINOiser due to the similarity of the approach. Nevertheless, ours varies in a smoother way, boosting the performance,  which indicates that our LP-based formulation is a key ingredient. 
Additionally, in Figure~\ref{fig:lposs_comparison}, we show the qualitative comparison of \ours and \oursplus. We observe that predictions of \oursplus follow the object boundary much better than the predictions from \ours.

\noindent\textbf{Impact of sliding windows.}
In Table~\ref{tab:window_impace}, we show the impact of sliding windows. All methods are negatively impacted when we depart from the use of sliding windows, with ProxyCLIP and LaVG being significantly more impacted than others. The largest drop in the performance happens on the Cityscapes dataset, which has an aspect ratio that is furthest from the square aspect ratio that was used to train VLM and VM. Besides that, CLIP-DINOiser and ProxyCLIP perform better when used with larger sliding windows, while LaVG, \ours, and \oursplus prefer smaller windows. We attribute \ours's and \oursplus's preference for smaller windows to the fact that we \emph{decouple} prediction and feature extraction. We are thus able to select a (smaller)
window size close to the training resolution to get stronger features, but are still able to propagate labels over the full image. Related methods lose performance in this setting because the prediction happens with the smaller context. Additionally, based on the analysis in Section~\ref{sec:failure_cases} of the supplementary we propose to ensemble the two different window sizes. We observe that all methods, besides LaVG, benefit from the ensemble. However, \ours and \oursplus benefit more than others and obtain state-of-the-art results.

\noindent\textbf{LPOSS++ as a refinement step for linear semantic segmentation.}
\oursplus refines pixel-level predictions and, as such, could potentially be applied to other dense prediction tasks. To test this, we apply it on the predictions obtained by training a linear segmentation head on top of DINOv2~\cite{oquab2024dinov2} features on the ADE20k dataset~\cite{zhou17scene,zhou19semantic}. Applying \oursplus out of the box, we are able to improve DINOv2 predictions and increase mIoU from $47.3\%$ to $49.5\%$.

\noindent\textbf{Comparison with a training-based approach.}
We compare \ours and \oursplus with a training-based method CAT-Seg~\cite{cho24catseg} in Section~\ref{sec:catseg_comparison}. We observe that \ours and \oursplus outperform CAT-Seg on datasets that are far from CAT-Seg's training distribution.

\section{Conclusion}
This work proposes a training-free method for open-vocabulary segmentation, called \ours, that uses ViT-based VLMs and VMs. 
We highlight the limitations of low-resolution patch-based predictions and window-based processing. 
\ours targets both issues and provides improvements through an LP-based predictor that operates jointly over patches of all windows, and over pixels.
In a comprehensive benchmark across a variety of datasets and domains, \ours surpasses the previous state-of-the-art, especially when evaluating near class boundaries.

\paragraph{Acknowledgements.} This work was supported by the Junior Star GACR GM 21-28830M and the Czech Technical University in Prague grant No. SGS23/173/OHK3/3T/13. We acknowledge VSB – Technical University of Ostrava, IT4Innovations National Supercomputing Center, Czech Republic, for awarding this project access to the LUMI supercomputer, owned by the EuroHPC Joint Undertaking, hosted by CSC (Finland) and the LUMI consortium through the Ministry of Education, Youth and Sports of the Czech Republic through the e-INFRA CZ (grant ID: 90254).

{
    \small
    \bibliographystyle{ieeenat_fullname}
    \bibliography{main}
}

\clearpage
\maketitlesupplementary

\begin{figure*}[!b]
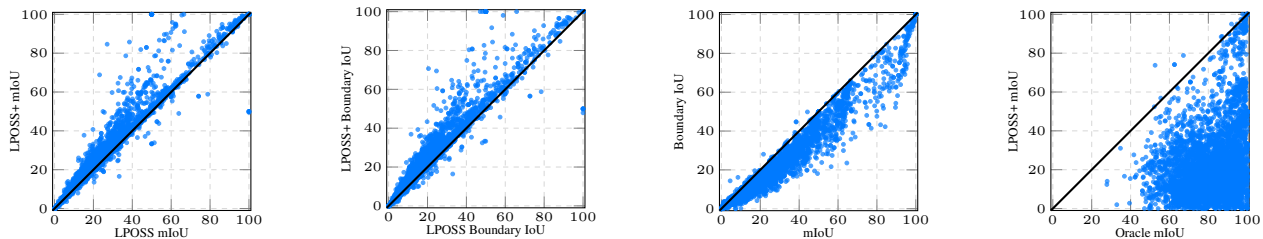

    \centering
    \begin{subfigure}[t]{0.24\linewidth}
        \centering
        \input{fig_supp/lposs_plus_lposs_miou}
        \caption{mIoU comparison of \ours and \oursplus}
        \label{fig:miou_lposs_lpossplus}
    \end{subfigure}
    \hfill
    \begin{subfigure}[t]{0.24\linewidth}
        \centering
        \input{fig_supp/lposs_plus_lposs_boundary}
        \caption{Boundary IoU comparison of \ours and \oursplus}
        \label{fig:boundary_lposs_lpossplus}
    \end{subfigure}
    \hfill
    \begin{subfigure}[t]{0.24\linewidth}
        \centering
        \input{fig_supp/miou_vs_boundary}
        \caption{Comparison between mIoU and Boundary IoU for \oursplus}
        \label{fig:miou_boundary}
    \end{subfigure}
    \hfill
    \begin{subfigure}[t]{0.24\linewidth}
        \centering
        \input{fig_supp/lposs_oracle_miou}
        \caption{mIoU comparison of Oracle (patch-level res.) and \ours}
        \label{fig:miou_oracle_lpossplus}
    \end{subfigure}
    \caption{\textbf{Analysis of \ours and \oursplus performance} per image. The plot shows 5000 randomly selected test images from all eight datasets used in the paper.}
\end{figure*}

\section{Analysis of performance}
\label{sec:failure_cases}

\subsection{Quantitative analysis per image}
In Figure~\ref{fig:miou_lposs_lpossplus} and Figure~\ref{fig:boundary_lposs_lpossplus}, we present the comparison of mIoU and Boundary IoU performance of \ours and \oursplus. We observe that \oursplus consistently improves the performance of \ours with respect to both metrics.

Figure~\ref{fig:miou_boundary} presents the comparison of mIoU and Boundary IoU performance of \oursplus. We see that although the two metrics are correlated to some extent, they are still complementary to each other. Boundary IoU performance is usually lower than mIoU performance, which is consistent with the definition that Boundary IoU measures the fine-grained performance at segment borders, which consist of the hardest pixels to classify.

We present the comparison of the mIoU performance of the oracle experiment and \ours in Figure~\ref{fig:miou_oracle_lpossplus}. We observe that oracle performance varies a lot across images, which we attribute to the differences in object size and shape. Additionally, the oracle performance acts as an upper bound in the majority of cases. Exceptions are justified by the effect of bilinear interpolation and combining predictions across many windows.

\subsection{Per image/class comparison with MaskCLIP}
\begin{figure}[t]
    {
\input{fig_supp/pgfplotsdata}
\definecolor{appleblue}{RGB}{0,122,255}
\pgfplotsset{every tick label/.append style={font=\tiny}}

\begin{tabular}{c@{\ssp}c@{\ssp}}
\includegraphics{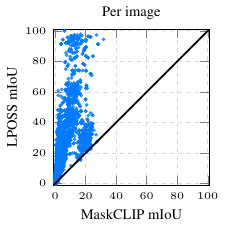}
&
\includegraphics{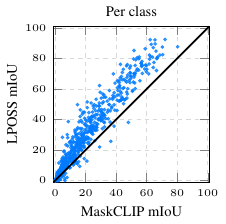}
\end{tabular}
}

    \caption{\textbf{Comparison of MaskCLIP and \ours performance} per image and per class. The plot shows 5000 randomly selected test images and all classes from all eight datasets in our experiments.}
    \label{fig:maskclip_lposs_scatter}
\end{figure}

\ours refines MaskCLIP predictions, so we look at how successful \ours is in improving these predictions. Figure~\ref{fig:maskclip_lposs_scatter} shows the comparison of the mIoU for \ours and MaskCLIP on the image and class level. We observe that \ours successfully improves MaskCLIP predictions in the vast majority of cases. Rarely, an image, with already very low performance, is further harmed if the MaskCLIP output is spatially very noisy, as shown in Figure~\ref{fig:maskclip_fail}.

\subsection{The impact of window size}
Furthermore, we look into the impact of window size on CLIP-DINOiser and \ours. In Figure~\ref{fig:dinoiser_fail}, we visualize the results of both methods as well as MaskCLIP using two different window sizes. We observe that there are some cases when MaskCLIP has better output for the large windows, used by CLIP-DINOiser, vs the small ones, used by \ours. In these cases, CLIP-DINOiser outperforms \ours. We also observe that there are cases where \ours performs well for both large and small window sizes, while CLIP-DINOiser performs better for the larger window size. Based on this, we propose to run our methods using an ensemble of large and small window sizes, and observe that this further improves the performance of \ours and \oursplus, as presented in Section~\ref{sec:results}.

{
\setlength{\tabcolsep}{-1.5pt}
\begin{figure}[b]
   \begin{center}
   \begin{tabular}{c@{\ssp}c@{\ssp}c@{\ssp}}
      Image(GT) & MaskCLIP & \ours \\ 
      \includegraphics[width=0.15\textwidth]{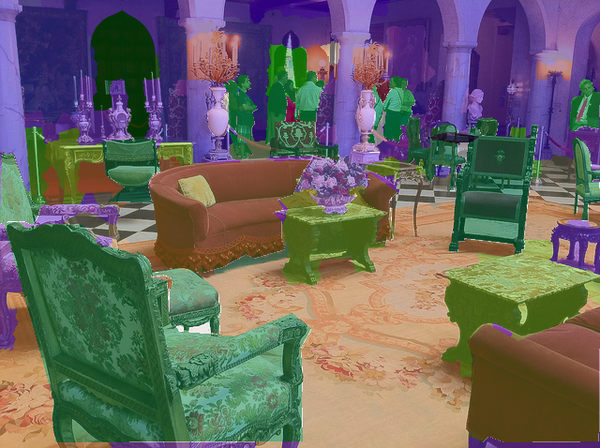}
      & 
      \includegraphics[width=0.15\textwidth]{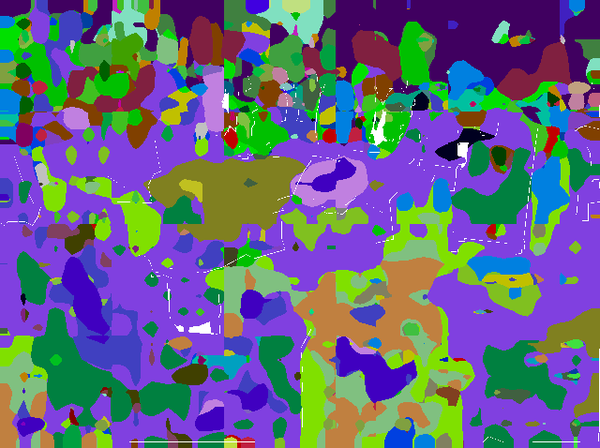}
      & 
      \includegraphics[width=0.15\textwidth]{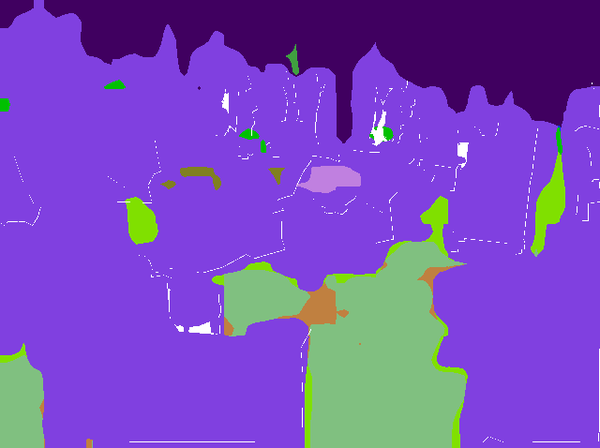}
      \\
      \includegraphics[width=0.15\textwidth]{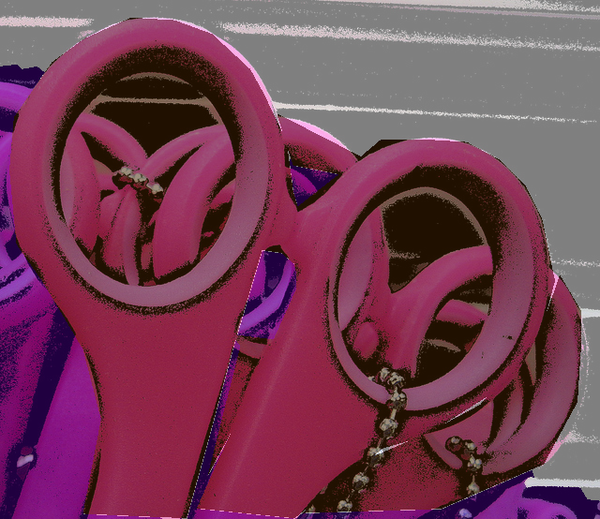}
      & 
      \includegraphics[width=0.15\textwidth]{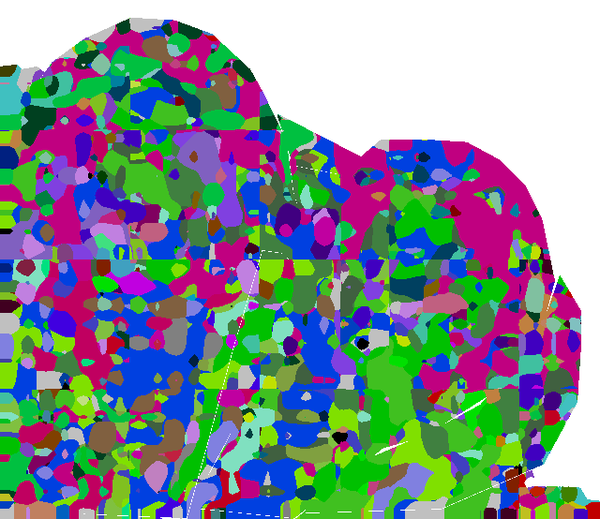}
      & 
      \includegraphics[width=0.15\textwidth]{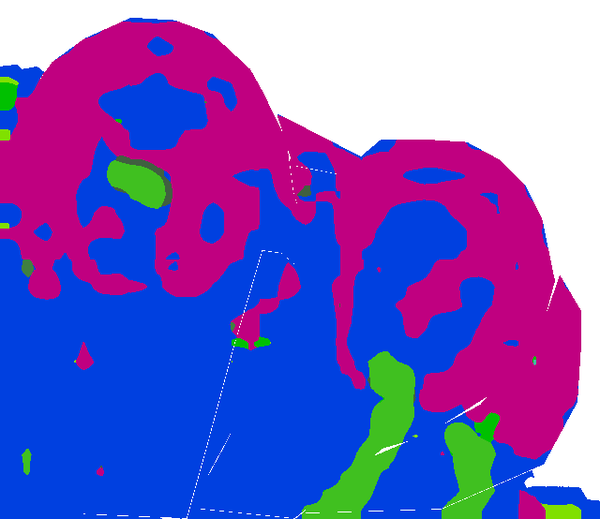}
      \\
      \includegraphics[width=0.15\textwidth]{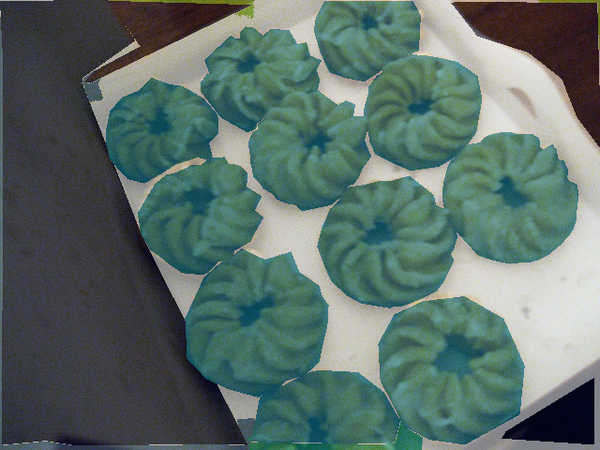}
      & 
      \includegraphics[width=0.15\textwidth]{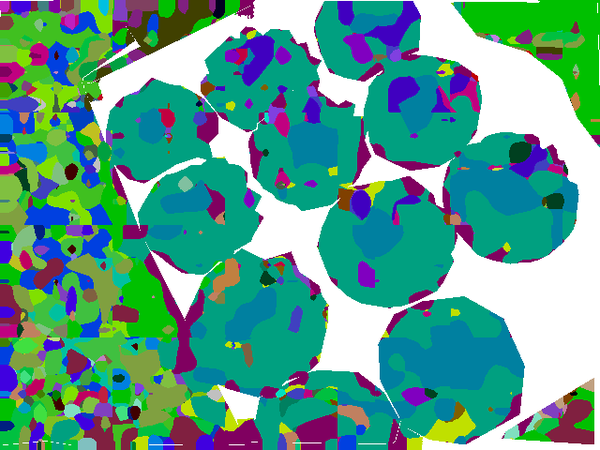}
      & 
      \includegraphics[width=0.15\textwidth]{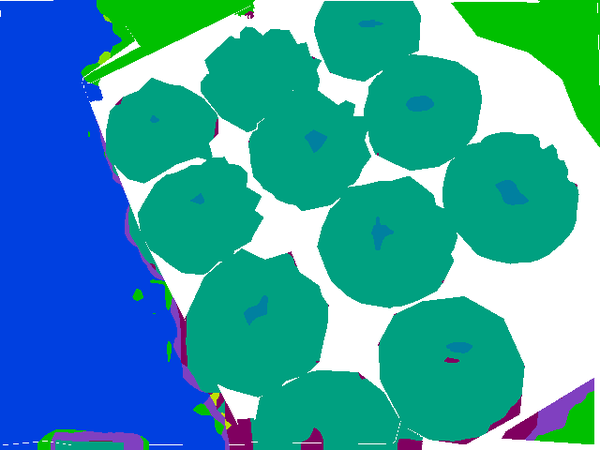}
      \\
      \includegraphics[width=0.15\textwidth]{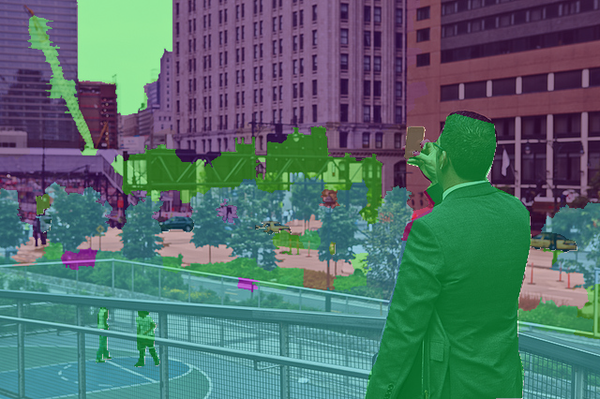}
      & 
      \includegraphics[width=0.15\textwidth]{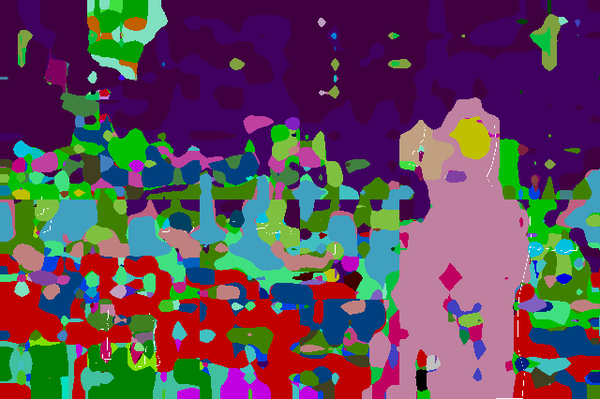}
      & 
      \includegraphics[width=0.15\textwidth]{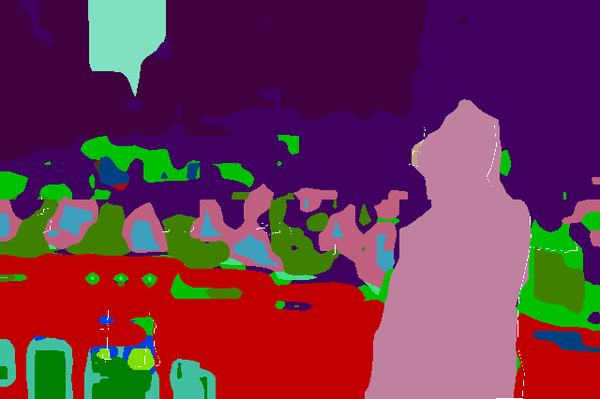}
   \end{tabular}
\end{center}
\vspace{-20pt}
\caption{\textbf{Examples of the \ours failure cases that are due to the MaskCLIP noisy predictions.} MaskCLIP produces very spatially noisy predictions in some cases, which then translates to the bad performance of \ours. Pixels shown in white are pixels that do not have a class in the ground-truth.
}
    \label{fig:maskclip_fail}
\end{figure}
}

{
\setlength{\tabcolsep}{-1.5pt}
\begin{figure*}[t]
\scriptsize
   \begin{center}
   \begin{tabular}{c@{\ssp}c@{\ssp}c@{\ssp}c@{\ssp}c@{\ssp}c@{\ssp}c@{\ssp}}
      Image(GT) & \textbf{MaskCLIP@448} & MaskCLIP@224 & \textbf{CLIP-DINOiser@448} & CLIP-DINOiser@224 & {\ours}@448 & \textbf{{\ours}@224} \\ 
      \includegraphics[width=0.135\textwidth]{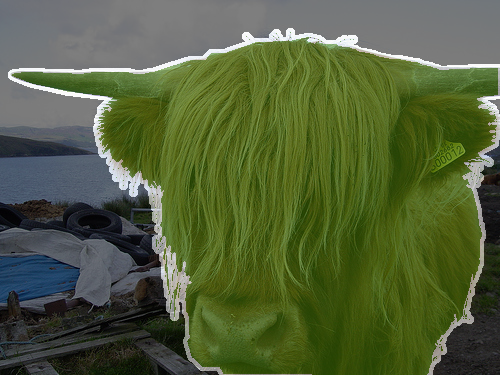}
      & 
      \includegraphics[width=0.135\textwidth]{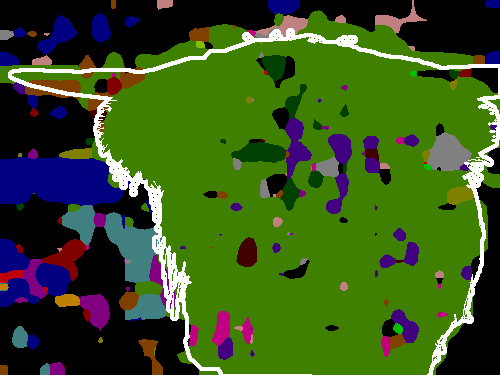}
      & 
      \includegraphics[width=0.135\textwidth]{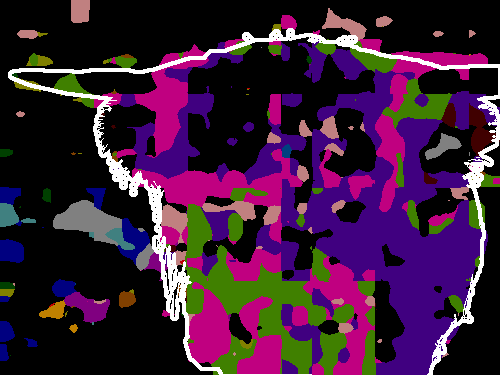}
      & 
      \includegraphics[width=0.135\textwidth]{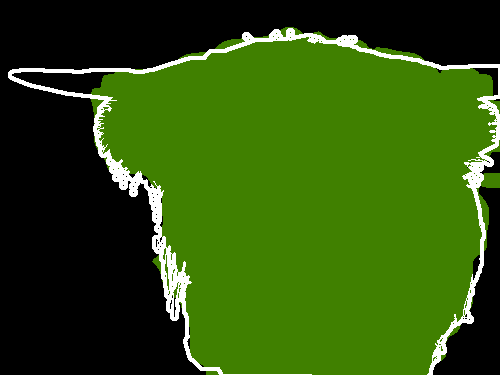}
      & 
      \includegraphics[width=0.135\textwidth]{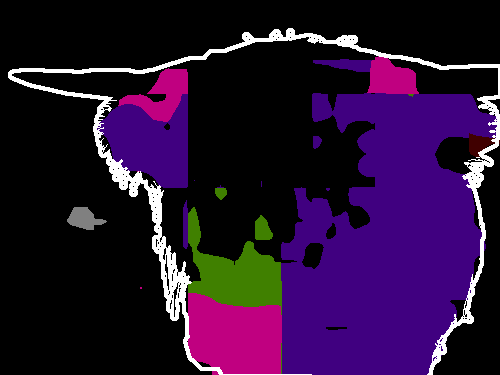}
      &
      \includegraphics[width=0.135\textwidth]{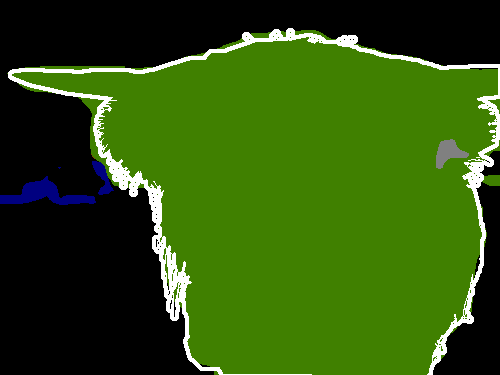}
      &
      \includegraphics[width=0.135\textwidth]{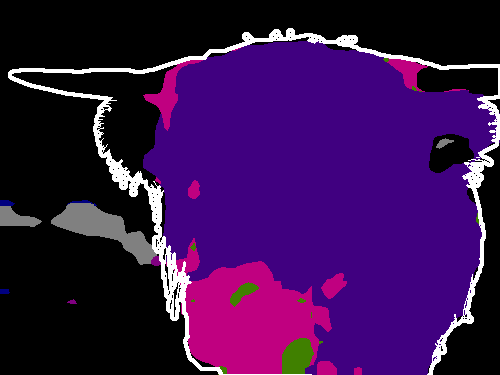}
      \\
      \includegraphics[width=0.135\textwidth]{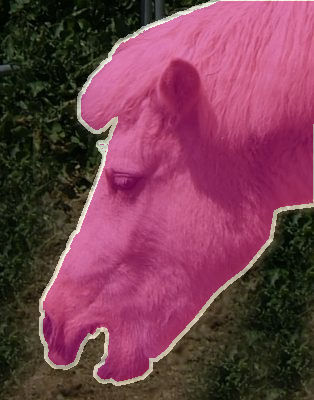}
      & 
      \includegraphics[width=0.135\textwidth]{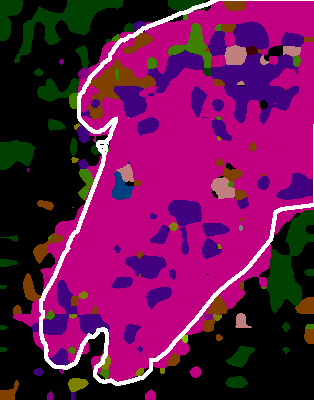}
      & 
      \includegraphics[width=0.135\textwidth]{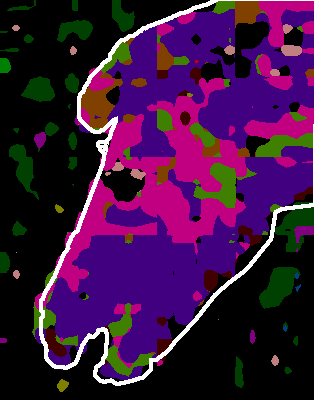}
      & 
      \includegraphics[width=0.135\textwidth]{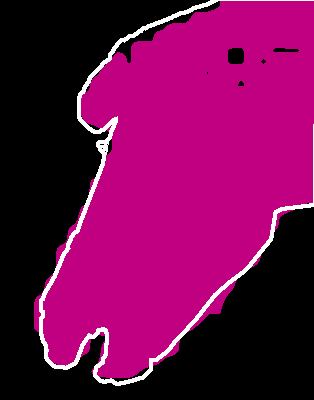}
      & 
      \includegraphics[width=0.135\textwidth]{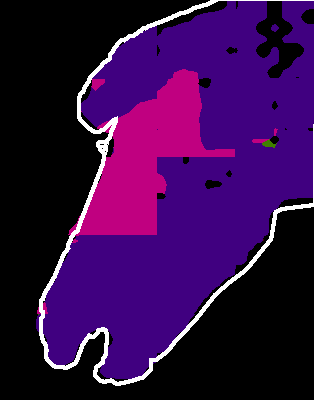}
      &
      \includegraphics[width=0.135\textwidth]{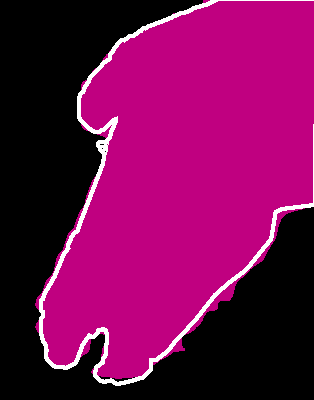}
      &
      \includegraphics[width=0.135\textwidth]{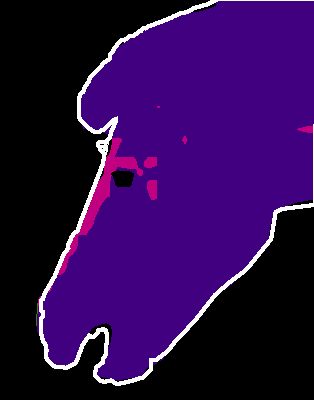}
      \\
      \midrule
      \includegraphics[width=0.135\textwidth]{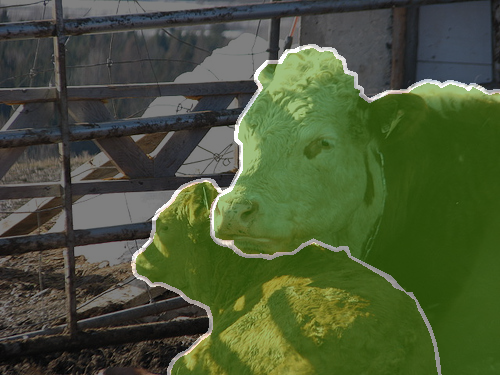}
      & 
      \includegraphics[width=0.135\textwidth]{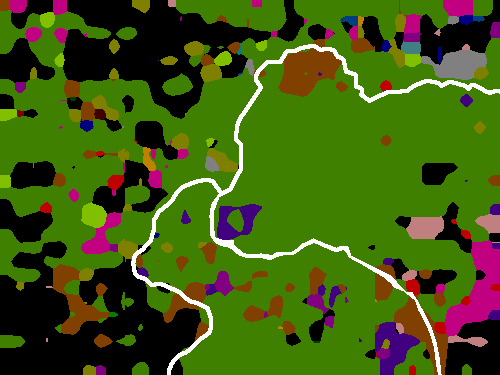}
      & 
      \includegraphics[width=0.135\textwidth]{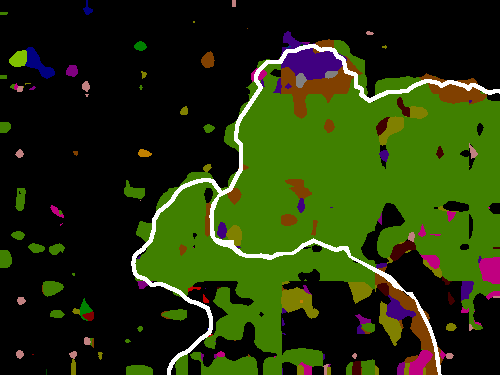}
      & 
      \includegraphics[width=0.135\textwidth]{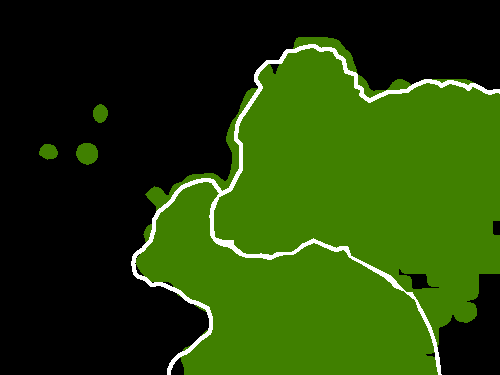}
      & 
      \includegraphics[width=0.135\textwidth]{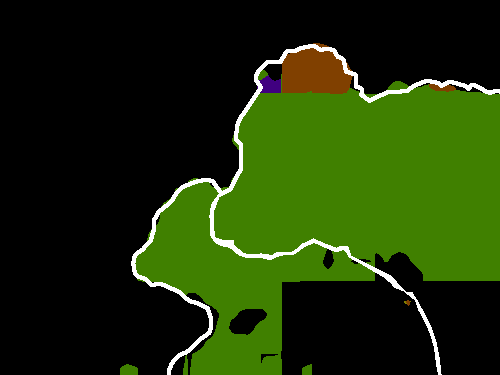}
      &
      \includegraphics[width=0.135\textwidth]{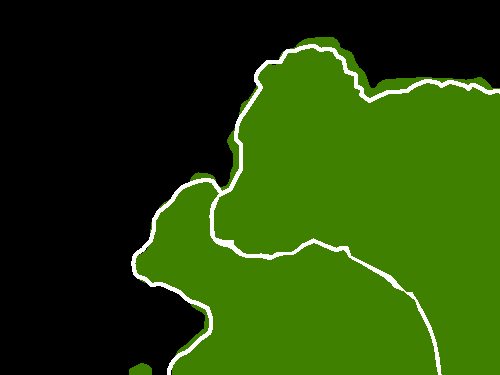}
      &
      \includegraphics[width=0.135\textwidth]{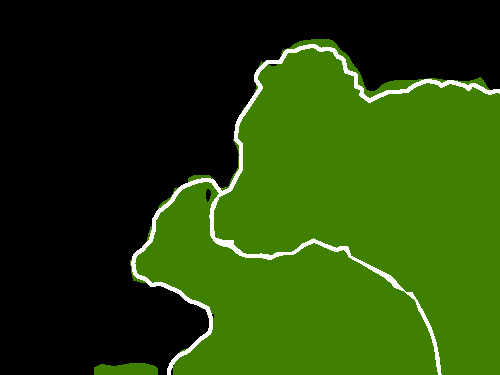}
      \\
      \includegraphics[width=0.135\textwidth]{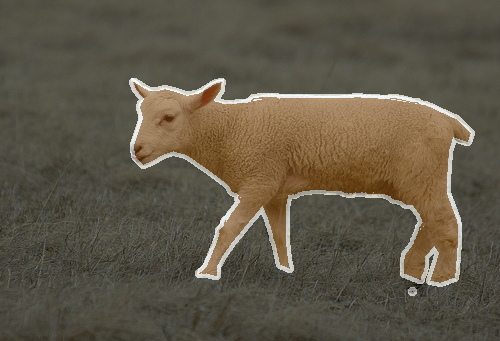}
      & 
      \includegraphics[width=0.135\textwidth]{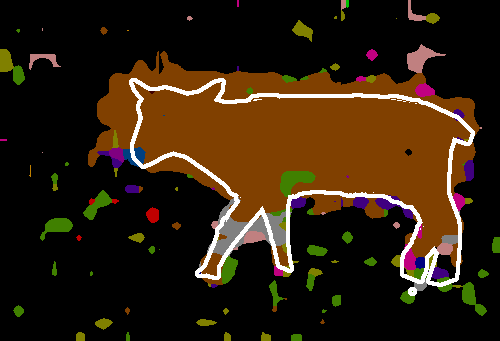}
      & 
      \includegraphics[width=0.135\textwidth]{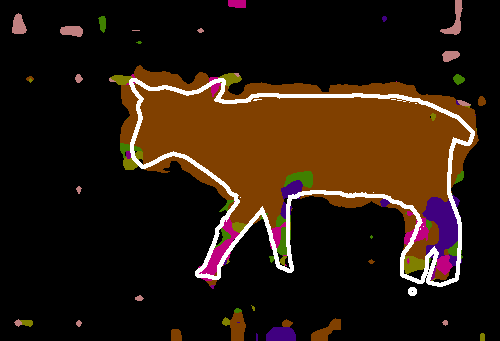}
      & 
      \includegraphics[width=0.135\textwidth]{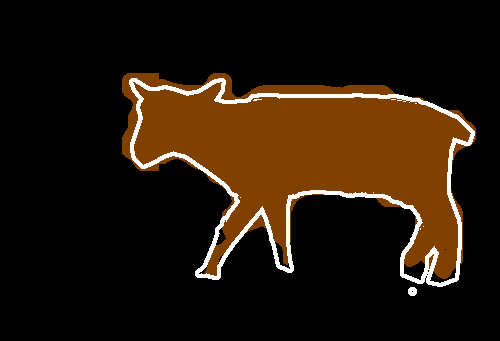}
      & 
      \includegraphics[width=0.135\textwidth]{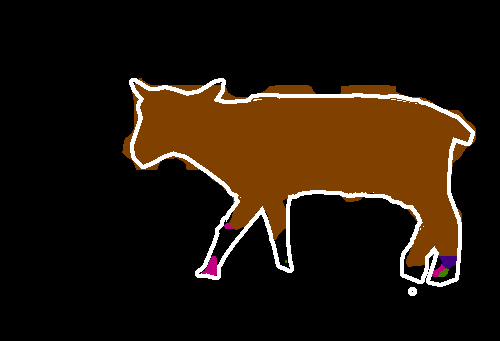}
      &
      \includegraphics[width=0.135\textwidth]{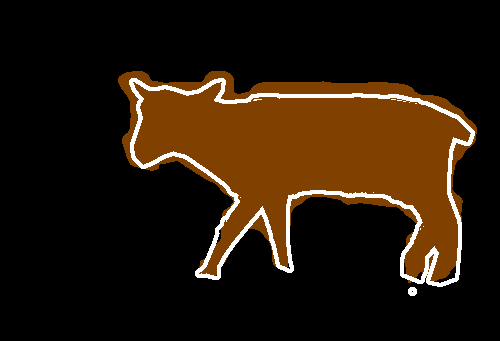}
      &
      \includegraphics[width=0.135\textwidth]{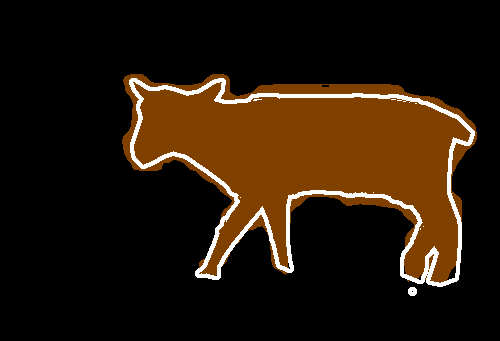}
   \end{tabular}
\end{center}
\vspace{-20pt}
\caption{\textbf{Impact of the window size on different methods.} The top rows show examples where both CLIP-DINOiser and \ours benefit from the large window size. The bottom rows show examples where \ours works well for both window sizes while CLIP-DINOiser performs better for the larger window size. Pixels shown in white are pixels that do not have a class in the ground-truth. Default setup of each method is shown in \textbf{bold}.
}
    \label{fig:dinoiser_fail}
\end{figure*}
}

\section{Ablation study}
\label{sec:desing_lposs}
\noindent\textbf{Design of $S_a$ and $S_p$.} We construct $S_a$ as a k-nearest neighbor graph with edge weights in the form $s^\gamma$, as it is a common choice for the adjacency matrix in the label propagation literature~\cite{iscen19lpsemi,stojnic24zlap}.
Another choice could be $\exp{(-\frac{1-s}{\sigma})}$, which we found to perform slightly worse for \ours ($-0.1\%$). For $S_p$, we chose the RBF kernel as in CRFs and bilateral filters. We further tested a linear kernel, which performs a bit worse ($-0.2\%$).

\ours and \oursplus construct adjacency matrices $S_{\mathtt{P}}$ and $S_{\tilde{\mathtt{P}}}$ in a different fashion. The primary difference comes from the way they control the sparsity of the graph. \ours uses k-nearest neighbors on top of the appearance-based adjacency $S_a$. On the other hand, \oursplus controls the sparsity using a binary spatial affinity $S_{\tilde{p}}$ that has non-zero elements only within the $r \times r$ neighborhood of the pixel. This difference is motivated by two reasons. First, with \oursplus, we aim to refine the predictions around the boundaries, which can be accomplished by looking at the neighborhood of each pixel. Second, the color-based features used in \oursplus can be very noisy, which can create issues for k-nearest neighbor search. To validate this, we try implementing \oursplus using the functions of \ours. However, we have found that this makes \oursplus ineffective and actually produces performance results worse than \ours. Additionally, we also experiment with using an RBF kernel in $S_{\tilde{p}}$ and find that it performs a bit worse than the proposed method (achieving the same performance as the proposed method when the RBF kernel converges to the proposed binary values). 

\noindent\textbf{Features used in adjacency $S_a$.}
We replace DINO features for the construction of $S_a$ with CLIP features and observe a significant drop in performance for both proposed methods, as shown in Table~\ref{tab:ablations}.

\noindent\textbf{Impact of spatial adjacency $S_p$.} We remove the use of $S_p$ in the construction of $S_{\mathtt{P}}$
, \ie set it to a matrix full of ones, and observe, as shown in Table~\ref{tab:ablations}, that using $S_p$ improves the results both for \ours and \oursplus. Additionally, in Figure~\ref{fig:spatial_ablation}, we visualize the impact of $S_p$ on the predictions. We observe that the use of $S_p$ cleans the predictions.

\noindent\textbf{Impact of window-based predictions.} We perform an experiment where we do feature extraction per window, as always, but also prediction per sliding window, as other methods do too~\cite{lan2024proxyclip,wysoczanska2024clip,kang2024defense}. We observe, as shown in Table~\ref{tab:ablations} that our choice of performing prediction jointly across all windows is indeed beneficial to the performance of \ours and \oursplus. 

\noindent\textbf{Pixel features.}
We switch pixel features from the default choice of Lab color space to depth predictions from DepthAnythingV2~\cite{yang24deptanythingv2} and their combination. We observe, as shown in Table~\ref{tab:ablations}, that using predicted depth marginally improves the results, but we opt not to use it in our default setup as it introduces another model during inference. We conclude that the use of different pixel-level features is worth future exploration.

\noindent\textbf{Impact of hyper-parameters $k$, $\sigma$, and $r$.} In Figure~\ref{fig:k_sigma_and_r}, we show the impact of hyper-parameters $k$, $\sigma$, and $r$ on the performance. Good performance is achieved over a
wide range of values. For \oursplus, $r$ controls the performance/speed trade-off (via sparsity), and we found $r = 13$ to be a good compromise.

{
\setlength{\tabcolsep}{-1.5pt}
\begin{figure*}[b]
   \begin{center}
   \begin{tabular}{c@{\ssp}c@{\ssp}c@{\ssp}c@{\ssp}c@{\ssp}}
      Image(GT) & MaskCLIP & \ours w/o $S_p$ & \ours w $S_p$ \\ 
      \includegraphics[width=0.2\textwidth]{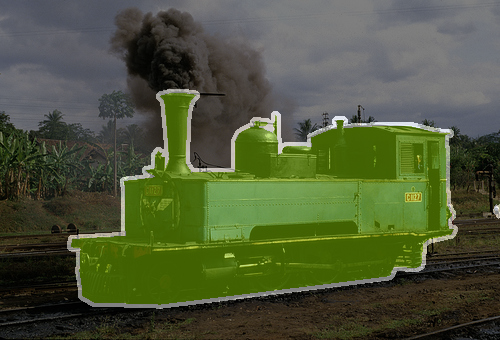}
      & 
      \includegraphics[width=0.2\textwidth]{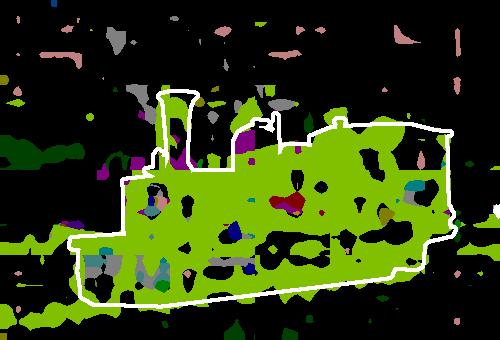}
      & 
      \includegraphics[width=0.2\textwidth]{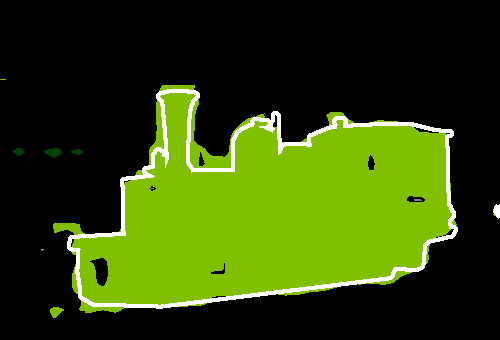}
      & 
      \includegraphics[width=0.2\textwidth]{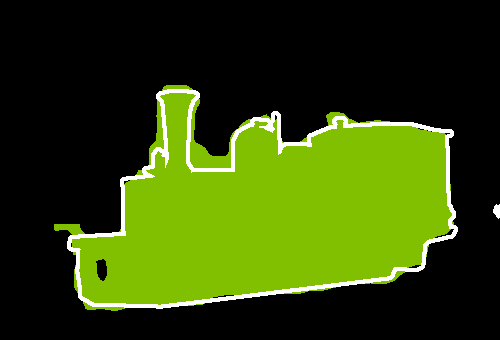}
      \\
      \includegraphics[width=0.2\textwidth]{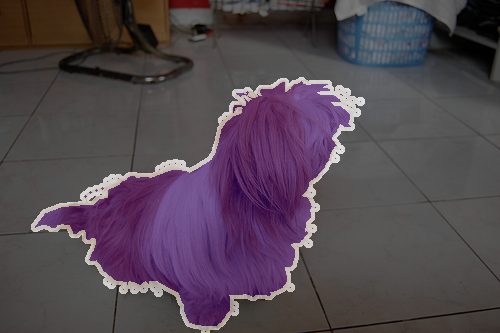}
      & 
      \includegraphics[width=0.2\textwidth]{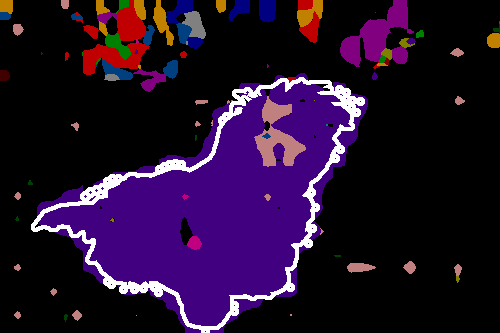}
      & 
      \includegraphics[width=0.2\textwidth]{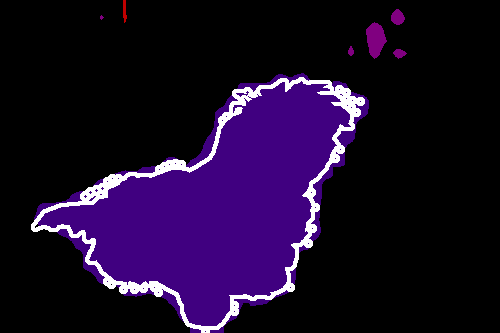}
      & 
      \includegraphics[width=0.2\textwidth]{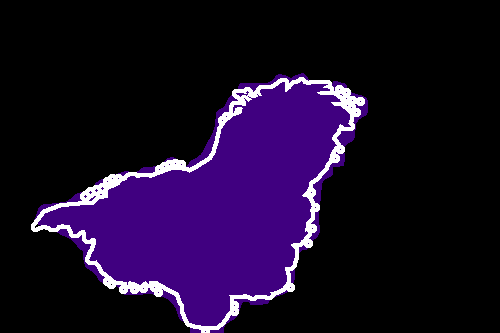}
      \\
      \includegraphics[width=0.2\textwidth]{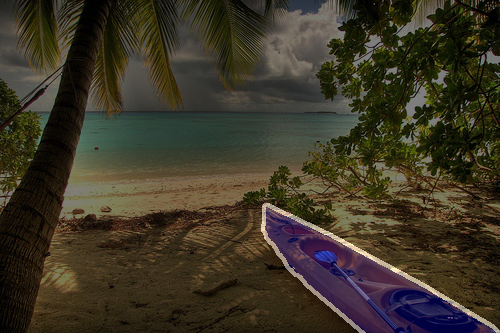}
      & 
      \includegraphics[width=0.2\textwidth]{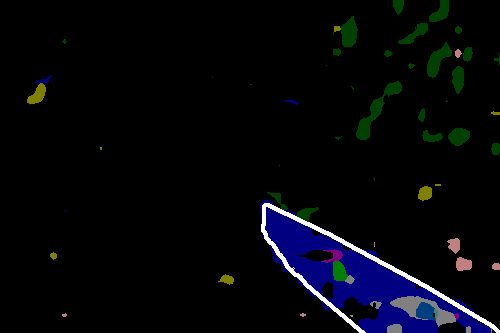}
      & 
      \includegraphics[width=0.2\textwidth]{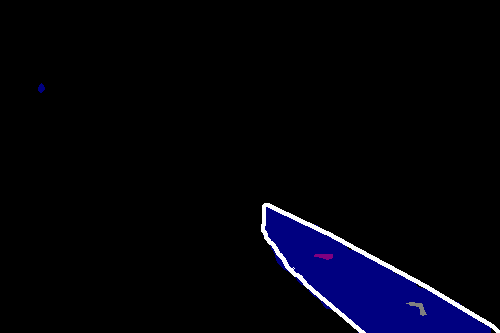}
      & 
      \includegraphics[width=0.2\textwidth]{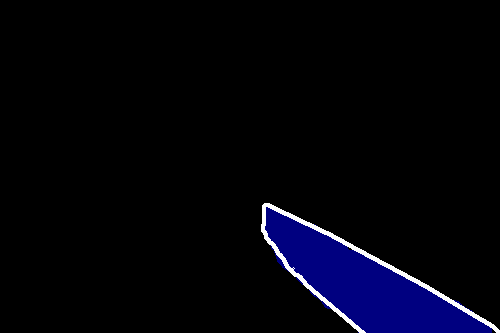}
      \\
      \includegraphics[width=0.2\textwidth]{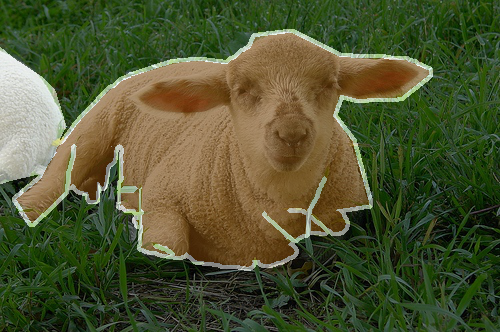}
      & 
      \includegraphics[width=0.2\textwidth]{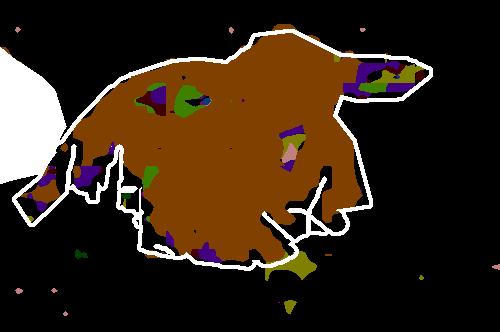}
      & 
      \includegraphics[width=0.2\textwidth]{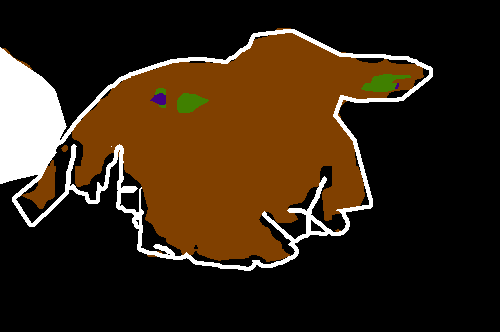}
      & 
      \includegraphics[width=0.2\textwidth]{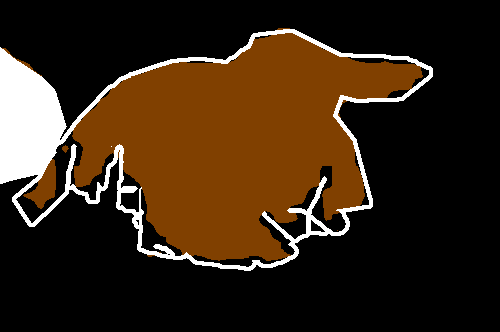}
   \end{tabular}
\end{center}
\vspace{-20pt}
\caption{\textbf{Impact of the spatial adjacency $S_p$.} A comparison of segmentation maps of \ours when applied without or with $S_p$. Pixels shown in white are pixels that do not have a class in the ground-truth.
}
    \label{fig:spatial_ablation}
\end{figure*}
}

\begin{table}[t]
    \centering
    {
\setlength{\tabcolsep}{2.5pt}
\newcommand{\cmark}{\ding{51}}%
\newcommand{\xmark}{\ding{55}}
\footnotesize
\begin{tabular}{lccccc}
\toprule
 &  &  & \multirow{3}{*}{\parbox{1.6cm}{Coupled feat. extraction and prediction}} &  &  \\
Method & $S_a$ & $S_p$ &  & Pix. feature & Avg \\
&  &  &  &  &  \\
\midrule
    \rowcolor{Gainsboro!60}
    \ours & DINO & \cmark & \xmark & - & \textbf{41.3} \\
    \ours & CLIP & \cmark & \xmark & - & 38.3 \\
    \ours & DINO & \xmark & \xmark & - & \underline{40.6} \\
    \ours & DINO & \cmark & \cmark & - & 39.2 \\
\midrule
    \rowcolor{Gainsboro!60}
    \oursplus & DINO & \cmark & \xmark & Lab & \underline{42.1} \\
    \oursplus & CLIP & \cmark & \xmark & Lab & 39.0 \\
    \oursplus & DINO & \xmark & \xmark & Lab & 41.4 \\
    \oursplus & DINO & \cmark & \cmark & Lab & 39.8 \\
    \oursplus & DINO & \cmark & \xmark & depth & \textbf{42.2} \\
    \oursplus & DINO & \cmark & \xmark & Lab+depth & \textbf{42.2} \\
\bottomrule
\end{tabular}
}

    \caption{\textbf{Ablations for \ours and \oursplus.} We report mIoU averaged across 8 datasets. Default setups used in the paper are marked with
    \colorbox{Gainsboro!60}{\phantom{aaa}}.}
    \label{tab:ablations}
\end{table}

\begin{figure*}[b]
  \centering
  \input{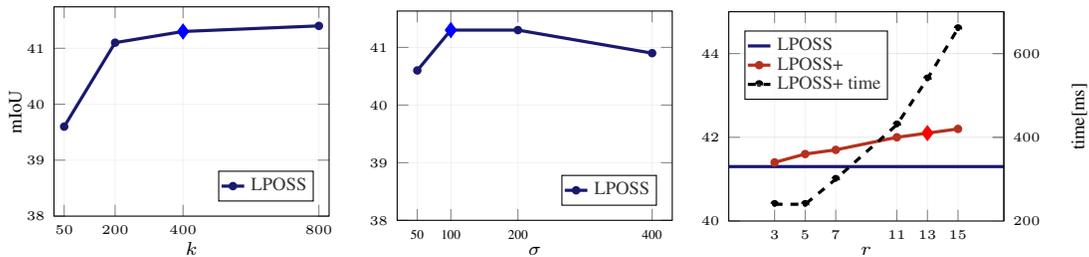}
\pgfplotsset{every tick label/.append style={font=\tiny}}
\pgfplotsset{select coords between index/.style 2 args={
    x filter/.code={
        \ifnum\coordindex<#1\def\pgfmathresult{}\fi
        \ifnum\coordindex>#2\def\pgfmathresult{}\fi
    }
}}
\pgfplotsset{minor grid style={solid,gray,opacity=0.1}}
\pgfplotsset{major grid style={solid,gray,opacity=0.1}}

\pgfplotsset{
compat=1.11,
legend image code/.code={
\draw[mark repeat=2,mark phase=2]
plot coordinates {
(0cm,0cm)
(0.15cm,0cm)        %
(0.3cm,0cm)         %
};%
}
}

{
    \tikzexternaldisable 
\setlength{\tabcolsep}{-1.5pt}
\begin{tabular}{c@{\msp}c@{\msp}c@{\msp}}
\tikzsetnextfilename{k_sigma_r_impact_1}
\begin{tikzpicture}
\begin{axis}[%
	width=0.3\linewidth,
	height=0.25\linewidth,
	xlabel={$k$},
	ylabel={mIoU},
	legend cell align={left},
    label style={font=\scriptsize},
	legend style={at={(0.62,0.10)},anchor=south west},
	legend style={cells={anchor=west}, font =\scriptsize, fill opacity=0.8, row sep=-3pt, inner sep=0.5pt, outer sep=-2.2pt},
	xmax = 830,
	xmin = 20,
    xtick={50, 200, 400, 800},
	ymin = 38,
    xlabel style = {yshift=4pt},
    ylabel style = {yshift=-5pt},
	grid=both,
]
    \addplot[color=MidnightBlue,     solid, mark=*,  mark size=1.0, line width=1.2] table[x=k, y expr={\thisrow{miou}}] \kimpact;\addlegendentry{\ours};
    \addplot [
        only marks,
        mark=diamond*,
        mark size=2.8pt,
        color=blue
    ] coordinates {
        (400,41.3)
    };
\end{axis}
\end{tikzpicture}
\tikzsetnextfilename{k_sigma_r_impact_2}
\begin{tikzpicture}
\begin{axis}[%
	width=0.3\linewidth,
	height=0.25\linewidth,
	xlabel={$\sigma$},
	legend cell align={left},
    label style={font=\scriptsize},
	legend style={at={(0.62,0.10)},anchor=south west},
	legend style={cells={anchor=west}, font =\scriptsize, fill opacity=0.8, row sep=-3pt, inner sep=0.5pt, outer sep=-2.2pt},
	xmax = 430,
	xmin = 20,
    xtick={50, 100, 200, 400},
    xticklabels={50,100,200,400},
	ymin = 38,
    xlabel style = {yshift=4pt},
    ylabel style = {yshift=-5pt},
	grid=both,
]
    \addplot[color=MidnightBlue,     solid, mark=*,  mark size=1.0, line width=1.2] table[x=s, y expr={\thisrow{miou}}] \sigmaimpact;\addlegendentry{\ours};
    \addplot [
        only marks,
        mark=diamond*,
        mark size=2.8pt,
        color=blue
    ] coordinates {
        (100,41.3)
    };
\end{axis}
\end{tikzpicture}
\tikzsetnextfilename{k_sigma_r_impact_3}
\begin{tikzpicture}
\begin{axis}[%
	width=0.3\linewidth,
	height=0.25\linewidth,
	xlabel={$r$},
	legend cell align={left},
        label style={font=\scriptsize},
    legend style={at={(0.1,0.88)},anchor=north west},
	legend style={cells={anchor=west}, font =\tiny, fill opacity=0.8, row sep=-3pt, inner sep=0.5pt, outer sep=-2.2pt},
	xmax = 18,
	xmin = 0,
    xtick={3, 5, 7, 11, 13, 15},
	ymin = 40, %
	ymax = 45,%
    xlabel style = {yshift=4pt},
    ylabel style = {yshift=-5pt},
	grid=both,
]
    \addplot[mark=none, MidnightBlue, line width=1.2, domain=0:18] {41.3};\label{lposs}
    \addplot[color=BrickRed,     solid, mark=*,  mark size=1.0, line width=1.2] table[x=r, y expr={\thisrow{miou}}] \rimpact;\label{lpossplus}
    \addplot [
        only marks,
        mark=diamond*,
        mark size=2.8pt,
        color=red
    ] coordinates {
        (13,42.1)
    };
\end{axis}

\begin{axis}[%
	width=0.3\linewidth,
	height=0.25\linewidth,
	xlabel={$r$},
	ylabel={time[ms]},
	legend cell align={left},
        label style={font=\scriptsize},
    legend style={at={(0.08,0.88)},anchor=north west},
	legend style={cells={anchor=west}, font =\scriptsize, fill opacity=0.8, row sep=-3pt, inner sep=0.5pt, outer sep=-2.2pt},
	xmax = 18,
	xmin = 0,
    xtick={3, 5, 7, 11, 13, 15},
	ymin = 200, %
	ymax = 700,%
    xlabel style = {yshift=4pt}, 
    ylabel style = {yshift=-5pt},
	grid=both,
    axis y line*=right,
  hide x axis,
]
    \addlegendimage{/pgfplots/refstyle=lposs}\addlegendentry{\ours}
    \addlegendimage{/pgfplots/refstyle=lpossplus}\addlegendentry{\oursplus}

    \addplot[color=black,     dashed, mark=*,  mark size=1.0, line width=1.2] table[x=r, y expr={\thisrow{time}}] \rimpact;\addlegendentry{\oursplus time};
\end{axis}
\end{tikzpicture}
\end{tabular}
\tikzexternalenable
}
   \caption{\textbf{Impact of hyper-parameters $k$, $\sigma$, and $r$.} We report mIoU averaged across 8 datasets. Default setups in the paper are marked with $\blacklozenge$.
}
   \label{fig:k_sigma_and_r}
\end{figure*}

\section{Additional results}
\subsection{Method comparison using a different VM.}
\looseness=-1 Throughout the paper, we use DINO~\cite{caron21dino} with ViT-B/16 backbone as a VM. However, here we show that \ours and \oursplus can be applied with other VM backbones as well. Concretely, we use DINO~\cite{caron21dino} with ViT-B/8 backbone and DINOv2~\cite{caron21dino} with ViT-B/14 backbone. Considering that the VM now uses a patch size different from that of a VLM, the feature vectors coming from the VM and VLM are of a different size. 
To apply \ours in such a situation, we upsample the VLM features 
to match the size of VM features.

We present the results of this experiment in Table~\ref{tab:vit_b_8} and compare the results of \ours and \oursplus with ProxyCLIP~\cite{lan2024proxyclip} and LaVG~\cite{kang2024defense}, which also report the performance of such experiments. 

\noindent\textbf{DINO with ViT-B/8.} For DINO with ViT-B/8, we observe that both \ours and \oursplus outperform LaVG, while ProxyCLIP outperforms \ours and performs on par with \oursplus. However, ProxyCLIP again significantly outperforms \ours and \oursplus only on Object and VOC20 datasets, for which we provide an explanation in Section~\ref{sec:results}. We also observe that going to pixel-level predictions in \oursplus improves the results even when the patch size is as small as 8.

Additionally, we observe that after using ViT-B/8 backbone for the VM, the graph defined by $S^{(\cW)}$ in~\equ{lposs_windows} has four times as many nodes compared to the default setup of using ViT-B/16 backbone, with the increase coming from the fact that there are four times more feature vectors for each sliding-window. So we propose to just increase the value of the hyper-parameter $k$ from 400 to 800, while keeping all other hyper-parameters fixed, to allow better connectivity between nodes coming from different sliding windows. With this change, we observe that \ours performs on par with ProxyCLIP, while \oursplus outperforms it, as shown in Table~\ref{tab:vit_b_8}.

\noindent\textbf{DINOv2 with ViT-B/14.} For DINOv2, we observe that \ours and \oursplus
perform slightly worse than for the case of using DINO with ViT-B/16. However, \oursplus still outperforms ProxyCLIP, while \ours performs on par with it. We note that due to the very different distribution of similarities coming from DINOv2, compared with DINO, we use a value of $\gamma = 7.0$ for \ours and \oursplus.

\begin{table*}[t]
    \centering
    
{
\footnotesize
\newcommand{\cmark}{\ding{51}}%
\newcommand{\xmark}{\ding{55}}
\begin{tabular}{lccccccccc@{\hspace{25pt}}c}
\toprule
Method & VM & VOC & Object & Context & C59 & Stuff & VOC20 & ADE20k & City & Avg \\
\midrule
    ProxyCLIP\textsuperscript{*}~\cite{lan2024proxyclip} & DINO(ViT-B/16) & 59.3 & \underline{36.3} & 34.4 & 38.0 & 25.7 & 79.7 & 19.4 & 36.0 & 41.1  \\
    LaVG\textsuperscript{*}~\cite{kang2024defense} & DINO(ViT-B/16) & 61.8 & 33.3 & 31.5 & 34.6 & 22.8 & \underline{81.9} & 14.8 & 25.0 & 38.2 \\\midrule
    \rowcolor{Gainsboro!60}
    \ours & DINO(ViT-B/16) & 61.1 & 33.4 & 34.6 & 37.8 & 25.9 & 78.8 & 21.8 & 37.3 & 41.3 \\
    \rowcolor{Gainsboro!60}
    \oursplus & DINO(ViT-B/16) & \underline{62.4} & 34.3 & 35.4 & 38.6 & \underline{26.5} & 79.3 & 22.3 & 37.9 & 42.1\\
\midrule
    ProxyCLIP~\cite{lan2024proxyclip} & DINO(ViT-B/8) & 61.3 & \textbf{37.5} & 35.3 & \textbf{39.1} & \underline{26.5} & 80.3 & 20.2 & 38.1 & \underline{42.3} \\
    LaVG~\cite{kang2024defense} & DINO(ViT-B/8) & 62.1 &	34.2 &	31.6 &	34.7 &	23.2 &	\textbf{82.5} &	15.8 &	26.2 &	38.8 \\\midrule
    \rowcolor{Gainsboro!60}
    \ours & DINO(ViT-B/8) & 61.4 &	33.5 &	34.9 &	38.2 &	26.3 &	77.6 &	22.6 &	\textbf{40.2} &	41.8 \\
    \ours ($k=800$) & DINO(ViT-B/8) & 62.2 &	34.1 &	35.2 &	38.5 &	26.4 &	78.7 &	22.5 &	\underline{39.4} &	42.1 \\
    \rowcolor{Gainsboro!60}
    \oursplus & DINO(ViT-B/8) & 62.2 &	34.2 &	\underline{35.5} &	\underline{38.9} &	\textbf{26.8} &	78.0 &	\textbf{23.0} &	\textbf{40.2} &	\underline{42.3}\\
    \oursplus ($k=800$) & DINO(ViT-B/8) & \textbf{63.0} & 34.8 &		\textbf{35.8} &	\textbf{39.1} & \textbf{26.8} &		79.0 &	\underline{22.8} &	39.3 &	\textbf{42.6}\\
\midrule
ProxyCLIP~\cite{lan2024proxyclip} & DINOv2(ViT-B/14) & 58.6 & \textbf{37.4} & 33.8 & 37.2 & 25.4 & \textbf{83.0} & 19.7 & 33.9 & \underline{41.1} \\
\midrule
\ours ($\gamma=7.0$) & DINOv2(ViT-B/14) & \underline{59.7} & 33.3 &		\underline{34.3} &	\underline{37.5} & \underline{25.6} &		80.0 &	\underline{21.9} &	\underline{36.0} &	41.0\\
\oursplus ($\gamma=7.0$) & DINOv2(ViT-B/14) & \textbf{60.8} & \underline{34.3} &		\textbf{35.1} &	\textbf{38.3} & \textbf{26.2} &		\underline{80.4} &	\textbf{22.4} &	\textbf{36.7} &	\textbf{41.8}\\
\bottomrule
\end{tabular}
}

    \caption{\textbf{Performance comparison in terms of mIoU on 8 datasets} using ViT-B/16 backbone for VLM and DINO ViT-B/8, DINO ViT-B/16, or DINOv2 ViT-B/14 backbone for VM. Default setups of hyper-parameters used in the paper are marked with \colorbox{Gainsboro!60}{\phantom{aaa}}. \textsuperscript{*} denotes the methods for which we reproduce the performance.}
    \label{tab:vit_b_8}
\end{table*}

\subsection{Complementarity with other methods.}
We build \ours and \oursplus by applying them on top of initial predictions coming from the VLM; in particular as MaskCLIP computes them. However, these initial predictions can come from any model, so here we show that we can apply them on top of CLIP-DINOiser~\cite{wysoczanska2024clip}, by simply using $Y_{\text{DINOiser}}$ instead of $Y_{\text{vlm}}$. Compared to the main paper experiments, we also use the sliding window setup used in CLIP-DINOiser, as it significantly improves CLIP-DINOiser performance as shown in Table~\ref{tab:window_impace}.

We present the results of these experiments in Table~\ref{tab:lposs_on_dinoiser}. We observe that \ours and \oursplus are complementary to CLIP-DINOiser and that they can further improve its performance. Additionally, we show that using a better initialization of CLIP-DINOiser compared to MaskCLIP improves \ours and \oursplus results as well.

\begin{table*}[t]
    \centering
    {
\footnotesize
\newcommand{\cmark}{\ding{51}}%
\newcommand{\xmark}{\ding{55}}
\begin{tabular}{lcccccccc@{\hspace{25pt}}c}
\toprule
Method & VOC & Object & Context & C59 & Stuff & VOC20 & ADE20k & City & Avg \\
\midrule
    MaskCLIP\textsuperscript{*}~\cite{zhou2022extract} & 32.9 & 16.3 & 22.9 & 25.5 & 17.5 & 61.8 & 14.2 & 25.0 & 27.0 \\
    \ours (MaskCLIP) & 61.1 & 32.5 & \underline{32.9} & 36.3 & 25.2 & 82.8 & \underline{20.4} & \underline{31.7} & 40.4 \\
    \oursplus (MaskCLIP) & 61.8 & 33.2 & \textbf{33.4} & \underline{36.9} & \textbf{25.6} & 83.1 & \textbf{20.7} & \textbf{31.9} & 40.8 \\
    \midrule
    CLIP-DINOiser\textsuperscript{*}~\cite{wysoczanska2024clip} & 62.2 & 34.7 & 32.5 & 36.0 & 24.6 & 80.8 & 20.1 & 31.1 & 40.2 \\
    \ours (CLIP-DINOiser) & \underline{65.0} &	\underline{36.3} &	\underline{32.9} &	36.6 &	25.1 &	\underline{84.0} &	19.7 &	29.3 &	\underline{41.1} \\
    \oursplus (CLIP-DINOiser) & \textbf{66.1} &	\textbf{36.8} &	\textbf{33.4} &	\textbf{37.2} &	\underline{25.4} &	\textbf{84.2} &	19.9 &	29.5 &	\textbf{41.6} \\
\bottomrule
\end{tabular}
}

    \caption{\textbf{Performance comparison in terms of mIoU on 8 datasets} applying \ours and \oursplus on top of MaskCLIP (default choice in the main paper) or CLIP-DINOiser~\cite{wysoczanska2024clip}. ViT-B/16 backbone is used both for VLM and VM. Sliding-window size $448 \times 448$ and stride $224 \times 224$ as used in CLIP-DINOiser. \textsuperscript{*} denotes the methods for which we reproduce the performance.}
    \label{tab:lposs_on_dinoiser}
\end{table*}

\subsection{\oursplus \vs other post-processing methods.}
We compare \oursplus with two other post-processing pixel refinement methods, PAMR~\cite{araslanov20pamr} and DenseCRF~\cite{krahenbuhl11densecrf} by applying them on top of \ours predictions. PAMR, with 5 iterations and dilations 8 and 16, achieves 41.8 mIoU (\vs 42.1 of \oursplus) while DenseCRF was unable to improve \ours at all.

\subsection{Comparison with training-based methods.}
\label{sec:catseg_comparison}
We compare variants of \ours and \oursplus with a training-based approach CAT-Seg~\cite{cho24catseg} on the MESS benchmark~\cite{blumenstiel23mess}. We present the results of these expemiments in Table~\ref{tab:mess_domains} and Table~\ref{tab:mess_all}, averaged across datasets of a domain and per dataset, respectively. We observe that \ours and \oursplus outperform CAT-Seg on domains that are far from the training domain of CAT-Seg, \ie Medical Sciences and Engineering, while CAT-Seg outperforms \ours and \oursplus on the General domain which is similar to CAT-Seg's training distribution. This shows that CAT-Seg does not generalize well to domains that are far from the training distribution. Additionally, CAT-Seg outperforms \ours and \oursplus on Earth Monitoring datasets, which we attribute to the fact that CAT-Seg uses larger resolution images, which allows it to better segment small objects, which are common in Earth Monitoring datasets.

\section{Computation requirements}

We measure the average time necessary to perform the inference per image on the VOC20 dataset using an NVIDIA A100 GPU. \ours processes each image in under 0.1s, which is comparable to all methods except LAVG (6.5s). \oursplus for pixel-level post-processing takes 0.5s/image, comparable to or faster than pixel-level post-processing methods PAMR~\cite{araslanov20pamr} (0.5s) or DenseCRF~\cite{krahenbuhl11densecrf} (1.6sec). Note that \oursplus speed can be controlled via hyper-parameter $r$ (see Figure~\ref{fig:k_sigma_and_r}).

\begin{table*}[t]
    \centering
    \resizebox{\textwidth}{!}{
\footnotesize
\begin{tabular}{lcccccccc}
\toprule
Method & VM & Sliding window & General & Earth Monitoring & Medical Sciences & Engineering & Biology & Non-general\\
\midrule
    CAT-Seg~\cite{cho24catseg} & - & - & \textbf{38.9}	& \textbf{36.1}	& 32.3	& 20.5	& \textbf{32.6} &  30.6 \\
\midrule
    \ours & DINO(ViT-B/16) & $224 \times 224$ & 29.6	& 28.6	& 41.9	& 28.4	& 30.8 & 32.3 \\
    \oursplus & DINO(ViT-B/16) & $224 \times 224$ & 30.3	& 29.1	& 41.8	& 28.8	& 31.3 & 32.6\\
    \ours & DINO(ViT-B/16) & ensemble & 29.9	& 28.7	& 44.0	& 29.5	& 30.7 & 33.1\\
    \oursplus & DINO(ViT-B/16) & ensemble & 30.5	& 29.0	& \underline{44.2}	& 29.8	& 31.1 & 33.4\\
\midrule
    \ours & DINO(ViT-B/8) & $224 \times 224$ & 30.9	& 29.5	& 42.3	& 28.7	& 31.0 & 32.8\\
    \oursplus & DINO(ViT-B/8) & $224 \times 224$ & 31.2	& \underline{29.8}	& 42.1	& 29.0	& 31.3 & 32.9\\
    \ours & DINO(ViT-B/8) & ensemble & 31.3	& 29.6	& \textbf{44.7}	& \underline{29.9}	& 31.2 & \underline{33.7}\\
    \oursplus & DINO(ViT-B/8) & ensemble & \underline{31.6}	& \underline{29.8}	& \textbf{44.7}	& \textbf{30.2}	& \underline{31.5} & \textbf{33.9}\\
\bottomrule
\end{tabular}
}

    \caption{\textbf{Performance comparison in terms of mIoU on different domains of MESS benchmark~\cite{blumenstiel23mess}}. ViT-B/16 backbone is used for VLM in \ours and \oursplus and for CAT-Seg~\cite{cho24catseg}. We vary the VM used for \ours and \oursplus. CAT-Seg~\cite{cho24catseg} does not employ sliding windows. \emph{Ensemble:} denotes combination of $224 \times 224$ and $448 \times 448$ sliding windows. \emph{Non-general:} average performance over datasets that do not belong to the general domain.}
    \label{tab:mess_domains}
\end{table*}

\begin{table*}[t]
    \centering
    \resizebox{\textwidth}{!}{
\footnotesize
\begin{tabular}{lcc|cccccc|ccccc|cccc|cccc|ccc|c}
\toprule
\multirow{2}{*}{Method} & \multirow{2}{*}{VM} & \multirow{2}{*}{Sliding window} & \multicolumn{6}{c}{General} & \multicolumn{5}{c}{Earth Monitoring} & \multicolumn{4}{c}{Medical Sciences} & \multicolumn{4}{c}{Engineering} & \multicolumn{3}{c}{Biology} & \\
 &  &  & \rotatebox{90}{BDD100K} & \rotatebox{90}{Dark Zurich} & \rotatebox{90}{MHPv1} & \rotatebox{90}{FoodSeg103} & \rotatebox{90}{ATLANTIS} & \rotatebox{90}{DRAM} & \rotatebox{90}{iSAID} & \rotatebox{90}{ISPRS Pots.} & \rotatebox{90}{WorldFloods} & \rotatebox{90}{FloodNet} & \rotatebox{90}{UAVid} & \rotatebox{90}{Kvasir-Inst.} & \rotatebox{90}{CHASE DB1} & \rotatebox{90}{CryoNuSeg} & \rotatebox{90}{PAXRay-4} & \rotatebox{90}{Corrosion CS} & \rotatebox{90}{DeepCrack} & \rotatebox{90}{PST900} & \rotatebox{90}{Zero Waste-f} & \rotatebox{90}{SUIM} & \rotatebox{90}{CUB-200} & \rotatebox{90}{CWFID} & \rotatebox{90}{Avg}\\
\midrule
    CAT-Seg~\cite{cho24catseg} & - & - & \textbf{47.0}	& \textbf{28.9}	& \underline{23.9}	& \textbf{26.7}	& \textbf{40.5}	& \textbf{66.3}	& \textbf{19.3}	& \textbf{45.4}	& 35.7	& \textbf{37.2}	& \textbf{43.0}	& 48.2	& 24.0	& 15.7	& 41.2	& 12.5	& 32.1	& \textbf{19.9}	& 17.5	& \textbf{44.8}	& 10.3	& \textbf{42.8}	& 32.9\\
\midrule
    \ours & DINO(ViT-B/16) & $224 \times 224$ & 36.2 & 17.6 & 20.9 & 25.6 & 33.3 & 44.1 & 13.0 & 25.4 & 37.7 & 28.9 & 38.2 & 61.1 & \underline{27.0} & 32.1 & \underline{47.2} & 13.6 & 55.6 & 19.6 & 24.7 & 39.7 & 11.7 & 41.1 & 31.6\\
    \oursplus & DINO(ViT-B/16) & $224 \times 224$ & 37.2 & 17.4 & 21.8 & 26.1 & 34.2 & 44.9 & 13.5 & 25.8 & 37.8 & 29.2 & 39.0 & 61.0 & 26.9 & 31.8 & \textbf{47.5} & 13.6 & 56.8 & 19.6 & \underline{25.1} & 40.3 & 12.0 & \underline{41.5} & 32.0\\
    \ours & DINO(ViT-B/16) & ensemble & 35.9 & 16.5 & 20.5 & 25.1 & 33.9 & 47.5 & 13.7 & 25.0 & 37.2 & 29.4 & 38.1 & 66.8 & \textbf{28.4} & 34.2 & 46.7 & 15.1 & 58.5 & 19.6 & 24.6 & 40.1 & \underline{13.5} & 38.6 & 32.2\\
    \oursplus & DINO(ViT-B/16) & ensemble &36.6 & 16.5 & 21.3 & 25.5 & 34.7 & 48.2 & 14.1 & 25.3 & 37.4 & 29.7 & 38.7 & 67.2 & \textbf{28.4} & 34.1 & 46.9 & 15.3 & \underline{59.6} & 19.5 & 24.8 & 40.6 & \textbf{13.8} & 39.0 & 32.6\\
\midrule
    \ours & DINO(ViT-B/8) & $224 \times 224$ & 38.1 & \underline{18.1} & 23.4 & 25.9 & 34.3 & 45.3 & 13.8 & \underline{27.2} & \underline{38.0} & 29.1 & \underline{39.6} & 62.9 & 25.0 & 33.8 & \textbf{47.5} & 14.5 & 55.6 & \underline{19.7} & 25.0 & 41.5 & 11.6 & 39.8 & 32.3\\
    \oursplus & DINO(ViT-B/8) & $224 \times 224$ & \underline{38.5} & 17.9 & \textbf{24.1} & \underline{26.2} & 35.0 & 45.7 & 14.3 & \underline{27.2} & \textbf{38.1} & 29.4 & 40.0 & 63.1 & 24.3 & 33.4 & \textbf{47.5} & 14.5 & 56.6 & 19.6 & \textbf{25.3} & \underline{41.8} & 11.7 & 40.3 & 32.5\\
    \ours & DINO(ViT-B/8) & ensemble & 37.6 & \underline{18.1} & 23.0 & 25.3 & 35.0 & 48.9 & 14.3 & 26.8 & 37.9 & 29.7 & 39.5 & \underline{68.4} & 26.3 & \underline{37.2} & 46.8 & \underline{15.8} & 59.2 & 19.6 & 24.9 & 41.4 & 13.4 & 38.7 & \underline{33.1}\\
    \oursplus & DINO(ViT-B/8) & ensemble & 37.9 & 17.9 & 23.6 & 25.6 & \underline{35.6} & \underline{49.2} & \underline{14.6} & 26.6 & \textbf{38.1} & \underline{29.9} & \underline{39.6} & \textbf{68.8} & 25.9 & \textbf{37.3} & 46.9 & \textbf{15.9} & \textbf{60.3} & 19.5 & \underline{25.1} & 41.6 & \underline{13.5} & 39.4 & \textbf{33.3}\\
\bottomrule
\end{tabular}
}

    \caption{\textbf{Performance comparison in terms of mIoU per dataset from MESS benchmark~\cite{blumenstiel23mess}}. ViT-B/16 backbone is used for VLM in \ours and \oursplus and for CAT-Seg~\cite{cho24catseg}. We vary the VM used for \ours and \oursplus. CAT-Seg~\cite{cho24catseg} does not employ sliding windows. \emph{Ensemble:} denotes combination of $224 \times 224$ and $448 \times 448$ sliding windows.}
    \label{tab:mess_all}
\end{table*}

\end{document}